\documentclass[preprint,12pt,numafflabel,sort&compress]{elsarticle}

\usepackage{amssymb}
\usepackage{amsmath}
\usepackage{lscape}
\usepackage{subfigure}
\usepackage{comment}
\usepackage{longtable}
\usepackage{booktabs}
\usepackage{multirow}
\usepackage{booktabs}
\usepackage{csquotes}
\usepackage{nomencl}
\usepackage{threeparttablex}
\usepackage{color}
\usepackage{xcolor}
\usepackage{hyperref}
\usepackage{soul}
\newtheorem{theorem}{Theorem}

\definecolor{myblue}{RGB}{0,0,0}
\usepackage[symbol]{footmisc}

\makenomenclature
\usepackage[pagewise]{lineno}
\setcitestyle{square}
\begin{document}

\begin{frontmatter}



\title{Robust Cross-Domain Adaptation in Texture Features Transferring for Wood Chip Moisture Content Prediction}
%Integrating Texture Features and Domain Adaptation for Robust Cross-Domain Moisture Content Prediction in Wood Chips


\author[insta]{Abdur Rahman}

\affiliation[insta]{organization={Department of Industrial Engineering},%Department and Organization
            addressline={Louisiana Tech University}, 
            city={Ruston},
            postcode={71270}, 
            state={LA},
            country={USA}}

\author[inst1]{Mohammad Marufuzzaman}
\affiliation[inst1]{organization={Department of Industrial and Systems Engineering},%Department and Organization
            addressline={Mississippi State University}, 
            city={MS State},
            postcode={39762}, 
            state={MS},
            country={USA}}
\author[inst2]{Jason Street}
\author[inst1]{Haifeng Wang\corref{cor1}}
\author[inst4]{Veera G. Gude}
\author[inst5]{Randy Buchanan}

\cortext[cor1]{Corresponding author: Haifeng Wang (E-mail: wang@ise.msstate.edu)}
\affiliation[inst2]{organization={Department of Sustainable Bioproducts},%Department and Organization
            addressline={Mississippi State University}, 
            city={MS State},
            postcode={39762}, 
            state={MS},
            country={USA}}

\affiliation[inst4]{organization={Purdue University Northwest Water Institute (PWI)},%Department and Organization
            addressline={Purdue University Northwest}, 
            city={Hammond},
            postcode={46323}, 
            state={IN},
            country={USA}}

\affiliation[inst5]{organization={US Army Engineer Research and Development Center},%Department and Organization
            addressline={3909 Halls Ferry Road},
            city={Vicksburg},
            postcode={39180}, 
            state={MS},
            country={USA}}
\begin{abstract}
Accurate and quick prediction of wood chip moisture content is critical for optimizing biofuel production and ensuring energy efficiency. The current widely used direct method (oven drying) is limited by its longer processing time and sample destructiveness. On the other hand, existing indirect methods, including near-infrared spectroscopy-based, electrical capacitance-based, and image-based approaches, are quick but not accurate when wood chips come from various sources. Variability in the source material can alter data distributions, undermining the performance of data-driven models. Therefore, there is a need for a robust approach that effectively mitigates the impact of source variability. Previous studies show that manually extracted texture features have the potential to predict wood chip moisture class. Building on this, in this study, we conduct a comprehensive analysis of five distinct texture feature types extracted from wood chip images to predict moisture content. Our findings reveal that a combined feature set incorporating all five texture features achieves an accuracy of 95\% and consistently outperforms individual texture features in predicting moisture content. To ensure robust moisture prediction, we propose a domain adaptation method named AdaptMoist that utilizes the texture features to transfer knowledge from one source of wood chip data to another, addressing variability across different domains. We also proposed a criterion for model saving based on adjusted mutual information and K-Means clustering methods to ensure the effective learning of the domain adaptation model. The AdaptMoist method improves prediction accuracy across domains by 23\%, achieving an average accuracy of 80\%, compared to 57\% for non-adapted models. These results highlight the effectiveness of AdaptMoist as a robust solution for wood chip moisture content estimation across domains, making it a potential solution for wood chip-reliant industries.
\end{abstract}

%\item Integration of various texture features improved moisture content prediction
%\item Model trained on one source of wood chip struggles to generalize on a new source of chips
%\item Adapt Moist effectively improved generalization capabilities for moisture prediction
%\item Custom AMI Callback is an innovative way to regulate domain adaptation model training



\end{frontmatter}


\section{Introduction}
\label{sec:introduction}

%why wood chip moisture content is important
%what are the existing ways to do it
%what are the limitations
    %did not explore multiple texture features
    %same source assumption
%our solutions
    %combination of textures
    %domain adaptation
%A challenge in domain adaptation is how much to adapt.
    %what are the existing solution to do this
    %what we proposed using Mutual Information driven callback.

Wood chips are widely utilized in biofuel production, making moisture content (MC) a critical parameter for optimizing energy efficiency and ensuring the quality of the final product. Accurately measuring the MC in wood chips is challenging due to variations in texture, environmental factors, and the complex nature of wood chip materials. Direct methods (e.g., oven drying) for predicting MC rely heavily on physical measurements, which are time-consuming and often inconsistent across different sources of wood chips \citep{rahman2024comprehensive}. Consequently, there has been growing interest in developing indirect data-driven approaches that are quick and non-invasive. In recent years, the utilization of machine learning and computer vision techniques has shown greater potential to enhance the accuracy and efficiency of MC prediction 
\citep{rahmaninterpretable, rahman2025moistnet, rahman2024moisture, gasperini2025machine}.

Indirect approaches utilize various characteristics of woodchips that have a correlation with MC under diverse data collection methods and then develop a data-driven approach to predict the moisture levels. These include techniques based on near-infrared (NIR) spectroscopy \citep{nascimbem2013determination, liang2019determination, amaral2020estimation, toscano2022performance, yan2024moisture}, electrical capacitance \citep{kandala2016capacitance, lev2021electrical, fridh2018precision, jensen2006moisture, pan2016predicting, de2023dielectric}, microwaves \citep{d2010simple, cazzorla2012woodchip, ottosson2018uwb}, Wi-Fi \citep{suthar2021multiclass}, X-rays \citep{kullenberg2010dual, hultnas2012determination, jain2017dual}, nuclear magnetic resonance (NMR) \citep{fridh2014accurate}, and image analysis \citep{plankenbuhler2020image, rahmaninterpretable, rahman2024moisture}. Among these, image-based approaches, particularly those utilizing neural networks, have shown significant promise for MC estimation \citep{rahmaninterpretable, rahman2025moistnet}.

NIR spectroscopy has been widely explored for MC determination. For instance, Nascimbem et al.~\cite{nascimbem2013determination} combined NIR spectroscopy with chemometric methods to evaluate quality parameters in moist wood chips, achieving a classification error below 6\% using partial least squares-discriminant analysis. They also developed robust calibration models using least squares support vector machines (LS-SVM). Similarly, Liang et al.~\cite{liang2019determination} applied NIR spectroscopy to predict MC in poplar wood chips under varying moisture conditions, while Amaral et al.~\cite{amaral2020estimation} conducted analogous studies on Eucalyptus wood chips. Toscano et al.~\cite{toscano2022performance} further evaluated the performance of portable NIR spectrometers for MC determination in wood chips. Despite its potential, NIR spectroscopy requires specialized equipment and is limited to surface-level measurements, with accuracy influenced by factors such as wood chip size distribution and geometry~\citep{liang2019determination, amaral2020estimation, toscano2022performance}.

Capacitance-based methods, leveraging the dielectric properties of wood chips, have also been developed for MC estimation. Kandala et al.~\cite{kandala2016capacitance} proposed a method for predicting MC in hardwood chips, noting higher accuracy for samples with MC below 25\%. Lev et al.~\cite{lev2021electrical} utilized an LCR meter to predict both MC and porosity, developing linear models through backward stepwise regression with \( R^2 \) values ranging from 0.9 to 0.99. However, Fridh et al.~\cite{fridh2018precision} found that the accuracy of handheld capacitance moisture meters decreased for wood chips with MC exceeding 50\%, highlighting a limitation of this approach. While capacitance-based methods effectively detect moisture variations, they assume wood chips to be a uniform material, which can lead to inaccuracies~\citep{pan2016predicting}.

Microwave-based techniques, such as the time-domain reflectometry (TDR) method proposed by D'Amico et al.~\cite{d2010simple}, measure wood-chip humidity by analyzing the Round Trip Time (RTT) of wire probe pulse signals. This approach has demonstrated sensitivity to humidity variations and the potential for low-cost monitoring systems. However, microwave methods are susceptible to inaccuracies due to air gaps between wood chips \citep{d2010simple}. Additionally, X-ray and Nuclear Magnetic Resonance (NMR) techniques, while effective, are costly and typically limited to small sample sizes \citep{barale2002use, hultnas2012determination}. The Wi-Fi-based method introduced by Suthar et al.~\cite{suthar2021multiclass} remains relatively unexplored but represents an emerging area of interest. Despite the widespread use of machine learning and statistical modeling, such as partial least squares (PLS) regression, in these indirect methods, the heterogeneous nature of wood chips complicates the accurate modeling of MC. This underscores the need for more advanced techniques capable of addressing these complexities and improving the robustness of MC determination methods.

One promising approach involves the extraction of texture features from wood chip images to predict MC \citep{rahman2024moisture}. Texture features capture the visual patterns and structural properties of the surface, providing valuable information for predicting moisture levels. However, previous studies have primarily focused on individual texture feature types, which may not fully exploit the potential of combined feature sets for enhanced prediction accuracy. In this study, we extracted five types of texture features, including Haralick \citep{haralick1973textural}, First-Order Statistics, Fourier Power Spectrum, Gray Level Run Length Matrix \citep{galloway1974texture}, and Local Binary Patterns \citep{ojala1996comparative}, from Red, Green, and Blue (RGB) images of wood chips collected under uniform lighting conditions. We employed these features to train and evaluate a set of twelve machine-learning classifiers to predict the moisture class of wood chips from three distinct sources. Additionally, we combined the five texture features to construct `combined features' and utilized them to evaluate the same models comprehensively. With this combination of texture features, we achieved state-of-the-art accuracy (95\% on Source 1) in predicting the moisture class of wood chips, outperforming MoistNet models \citep{rahman2025moistnet} and the Haralick texture-based study \citep{rahman2024moisture}. 

While developing any data-driven moisture classifier, existing studies \citep{rahman2025moistnet, rahman2024moisture} assumed that the training and test data come from the same distribution, which is commonly referred to as the independent and identically distributed (i.i.d.) assumption. However, wood chips are extremely heterogeneous materials. The heterogeneity of wood chips could originate from the shape, color, species of the plant, cutting method, harvesting time, and geographical location \citep{rahman2024comprehensive}. Heterogeneity leads to a shift in the distribution of datasets generated from wood chips. This violates the i.i.d. assumption of the existing data-driven moisture prediction frameworks. In this study, we relaxed this i.i.d. assumption of the training and test data. We collected wood chips from three different sources (e.g., lumber mills and forest environments). We conducted numerical experiments to check if the trained machine learning models can generalize when tested on a new source of data. The degraded performance of ML models due to distribution shifts necessitates a robust framework that generalizes well.

In this study, we proposed a robust framework based on the domain adversarial neural network \citep{ganin2016domain} to minimize this distribution shift in both training and test data. The proposed domain adaptation framework, named AdaptMoist, consists of three component networks: a feature extractor, a domain discriminator, and a label classifier. The feature extractor functions as a generator network that extracts domain-invariant features with the assistance of the gradient reversal layer from the domain discriminator. These domain-invariant features are then utilized to transfer the learned knowledge from one domain to a new domain. Since we do not have label information for the new wood chips, this results in an unsupervised domain adaptation framework. 

%mutual information driven callback
In unsupervised domain adaptation, evaluating the trained model is one of the fundamental challenges due to the lack of target domain labels. Although the source domain can be divided into training and validation sets, this validation does not provide an accurate validation score. The underlying distributions of the training and validation sets are essentially the same, failing to represent the domain adaptation scenario. This is why properly training a domain adaptation model is challenging. 

In unsupervised domain adaptation, discriminability focuses on maintaining the model’s ability to differentiate between classes within a specific domain, while transferability aims to enable the model to generalize across domains by learning domain-invariant features. If training continues for too long, the model may overly concentrate on aligning features between domains to improve transferability, potentially sacrificing the detailed class-specific features essential for discriminability. Conversely, if training is halted too early, the model may not have sufficient time to effectively align domains, which could harm transferability by leaving the model excessively specialized to the source domain. To balance these competing goals, this study proposes an Adjusted Mutual Information (AMI)-based callback to stop training at the optimal point, ensuring the model retains both adequate discriminability within the source domain and transferability to the target domain. According to Theorem \ref{th:1}, demonstrated by Ben David \citep{ben2010theory}, sufficient discriminability in the source domain (minimum source domain error, $\, \epsilon _s(h)$) and transferability to the target domain (minimum divergence between source and target domains, $d_1(\mathcal{D}_s,\mathcal{D}_t)$), provides a tighter bound on the target domain error, $\, \epsilon_t(h)$. 

\begin{theorem}
\label{th:1}
    For a hypothesis $h\in \mathcal{H}$ and two domain $\mathcal{D}_s$ and $\mathcal{D}_t$, we have
   \begin{equation*}
        \begin{split}
        \epsilon_t(h) \leq \epsilon_s(h) & + d_1(\mathcal{D}_s,\mathcal{D}_t) \\
        & + \min\{\mathrm{E}_{\mathcal{D}_s}[| f_s(\mathbf{x}) - f_t(\mathbf{x}) |], \\
        & \quad \mathrm{E}_{\mathcal{D}_t}[| f_s(\mathbf{x}) - f_t(\mathbf{x}) |]\}
        \end{split}
    \end{equation*}
    where $\epsilon _t(h)$ and $\epsilon _s(h)$ represent the error for target and source domains, respectively.
    The third term on the right side is the difference between the labeling functions across the two domains, which is expected to be negligible. The second term represents the $L1$ divergence measure between two domains. 
\end{theorem}

In this study, AMI serves as a model-saving (callback) criterion to monitor the alignment of clustering between the predicted target domain labels and the pseudo-labels generated for the target domain using a clustering algorithm. We assumed that the target data could be represented in clusters through methods such as K-Means clustering. This approach offers an efficient way to halt model training and save the best weights based on AMI. 

%Fuzzy CMeans Clustering \citep{bezdek2013pattern}. Fuzzy C Means is a soft clustering technique allowing for probabilistic cluster assignments, contrasting with the exclusive assignments of hard clustering algorithms like K-Means. To avoid the assignment of samples in wrong clusters, we have used d\% of the target domain data on which the clustering model is highly confident. In this way, we proposed an efficient way of stopping the model training and saving the best weight based on AMI. 

The contribution of this study can be summarized as:
\begin{enumerate}
    \item A comprehensive analysis of texture features (five types of individual features and their combinations) for predicting wood chip moisture content.
    \item For the first time, we relaxed the i.i.d. assumption regarding wood chip moisture level prediction data and tested whether existing methods can be generalized.
    \item To adapt to new sources of wood chip data, we proposed an adversarial domain adaptation method that enhances model generalization.
    \item To ensure the proper learning of the domain adaptation model, we proposed a model-saving callback based on adjusted mutual information and K-Means clustering methods. 
\end{enumerate}
The remainder of this paper is organized as follows: Section 2 outlines the methodology employed for feature extraction, the proposed AdaptMoist method, and the AMI callback. Section 3 presents the results of our experiments, which is followed by a discussion of the implications of our findings. Finally, Section 4 concludes the paper with suggestions for future research.
%Section 2 discusses the current method for predicting the MC of wood chips and provides an overview of domain adaptation approaches.
\section{Material and Methods}
\label{sec:method}

\subsection{Wood Chip Dataset Acquisition}
\begin{figure}[h]
    \centering
    \includegraphics[width=0.6\linewidth]{ 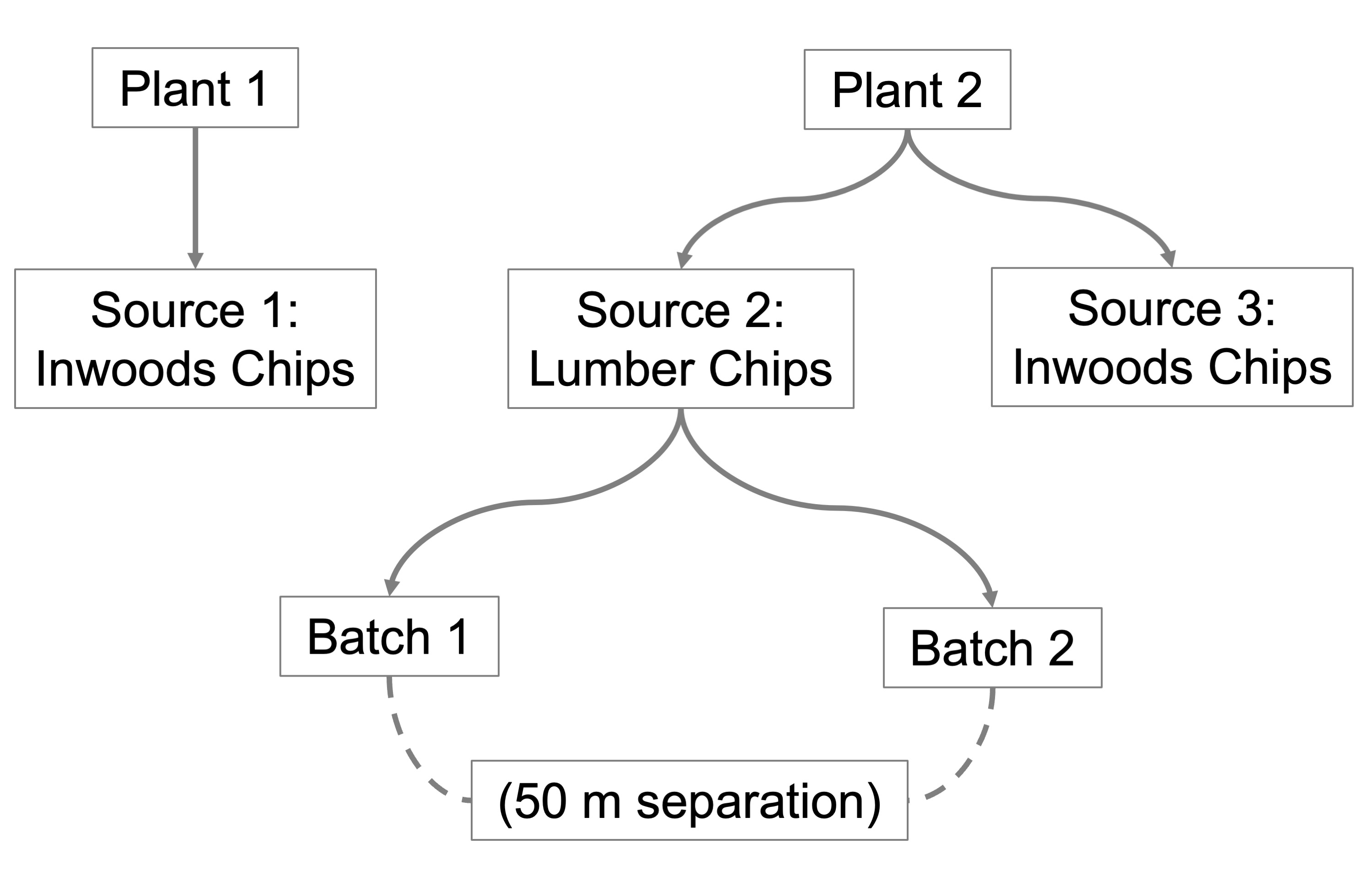}
    \caption{Wood chip acquisition method from multiple sources.}
    \label{fig:data_source}
\end{figure}

The wood chip dataset used in this study was initially presented and detailed in Rahman et al.~\cite{rahman2025moistnet}. Wood chips were obtained from two biofuel wood pellet processing plants designated as Plant 1 and Plant 2. Wood chips were stored in large piles, and a sampling strategy was employed following the procedures outlined in ASTM D6883 (61 cm depth). Chips collected directly from a forest environment were designated as inwoods chips. Chips sourced from kiln-dried lumber end-cuts were designated as lumber chips. The designations of the chips used in this study are as follows: Plant 1, inwoods chips (Source 1), Plant 2, lumber chips (Source 2), and Plant 2, inwoods chips (Source 3). To ensure robustness in MC prediction, Plant 2, Source 2 was further subdivided by sampling from two different points spaced approximately 50 meters apart, designated as Batch 1 (B1) and Batch 2 (B2). The wood chip acquisition method has been outlined in Figure \ref{fig:data_source}.

The dataset preparation involved artificially adjusting the MC of the wood chips to ensure a wide range of moisture levels suitable for deep-learning model training. Specifically, the chips were first oven-dried at 105°C for 24 hours and then rehydrated by adding measured amounts of water to achieve target MC levels. This process was rigorously controlled to account for moisture loss during mixing, and multiple moisture levels were generated, ensuring diverse data representation.

\begin{figure}[h]
    \centering
    \includegraphics[width=0.7\linewidth]{ 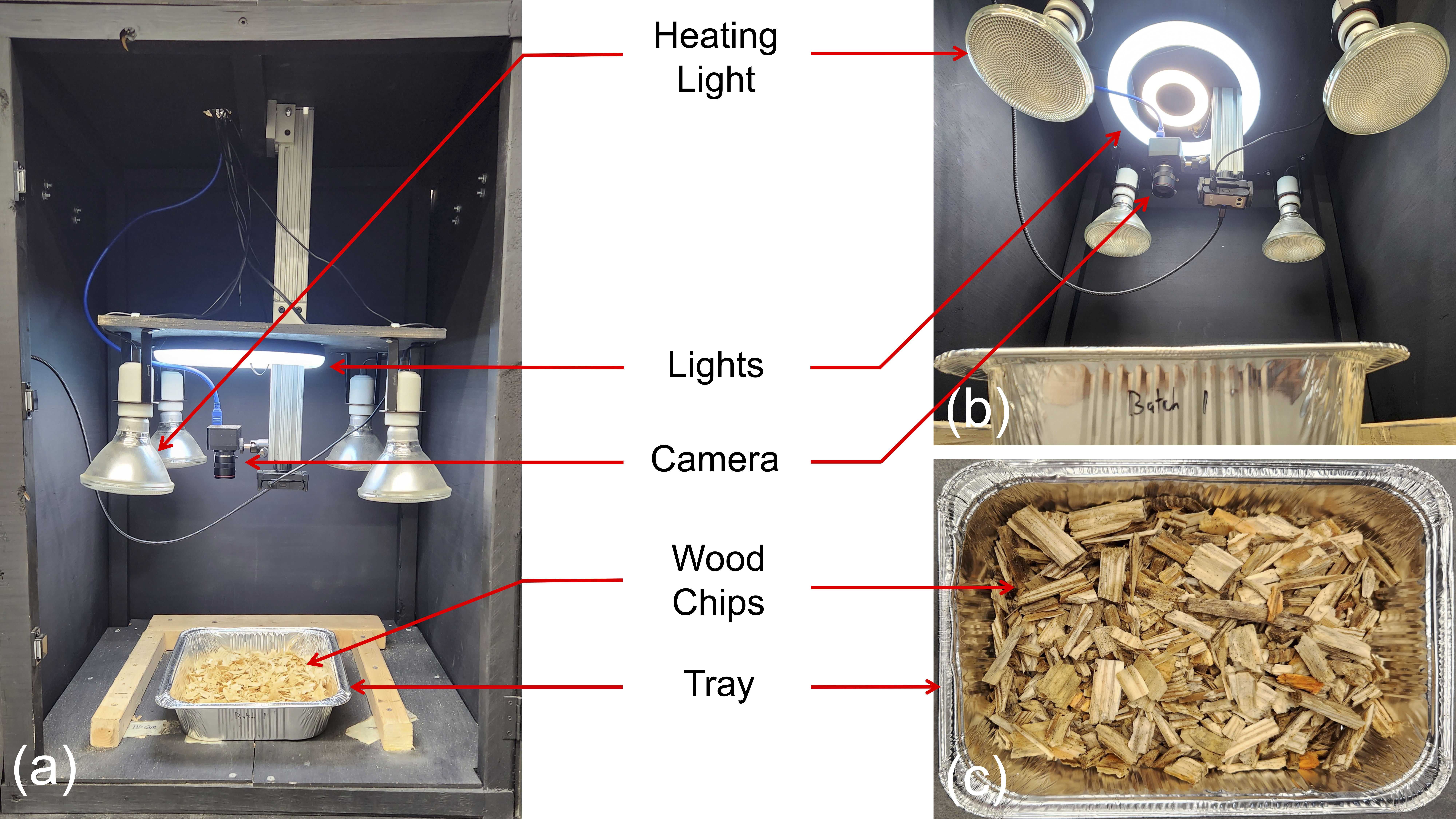}
    \caption{Wood chip data collection station: (a) Data collection setup, (b) Uniform white lights, (c) Wood chips in the container to capture images (Reproduced from Rahman et al.~\cite{rahman2025moistnet} with permission).}
    \label{fig:station}
\end{figure}
For image acquisition, wood chip samples were arranged in containers and placed in a controlled lighting environment to ensure uniform image quality, as shown in Figure \ref{fig:station}. An industrial camera (Hotpet 8MP USB Industrial Camera with Sony IMX179 Sensor) was used to capture RGB images of the chips from multiple angles by shuffling the chips after each capture to expose different surfaces. This method allowed for a comprehensive capture of moisture-related features beyond the surface-level moisture commonly captured by previous methods.

\begin{figure}[h]
    \centering
    \includegraphics[width=0.7\linewidth]{ 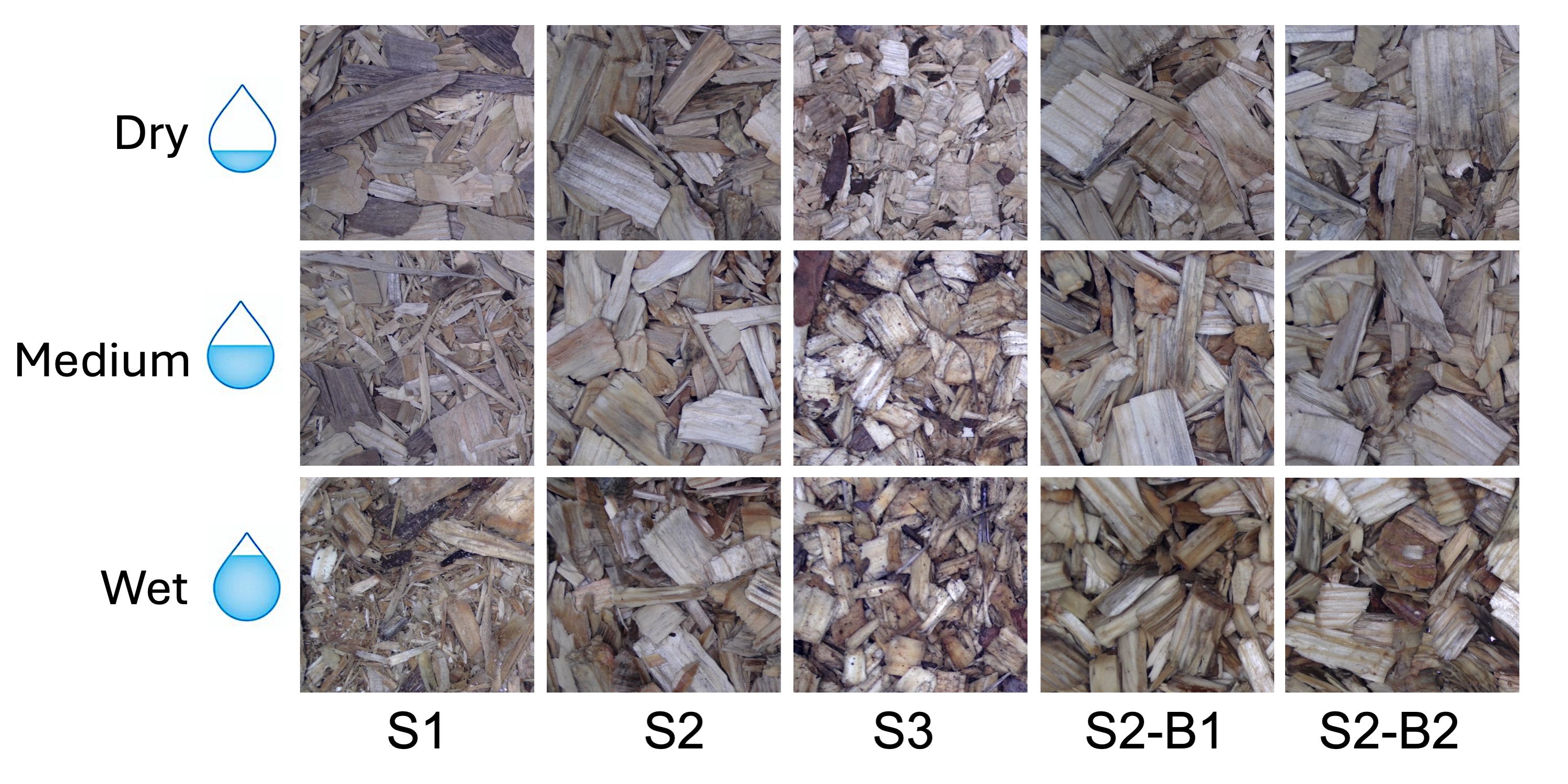}
    \caption{Sample images from the wood chip dataset. S1, S2, and S3 correspond to three different sources, and S2-B1 and S2-B2 represent two batches from S2. Dry, medium, and wet designations represent the three moisture content ranges.}
    \label{fig:dataset}
\end{figure}

The dataset was labeled using the oven-drying method, where the weight difference before and after drying was used to calculate the actual MC. The MC levels were subsequently categorized into three classes: \textit{dry} ($\leq$15\%), \textit{medium} (15\%$<$MC$<$35\%), and \textit{wet} ($\geq$35\%), based on discussions with industry partners. Full details regarding the dataset acquisition process can be found in Rahman et al.~\cite{rahman2025moistnet}. Figure \ref{fig:dataset} illustrates the sample images collected from each source and batch of wood chips.  S1, S2, and S3 include 800, 800, and 400 wood chip images, respectively. S2 has been divided into two batches, where S2B1 and S2B2 contain 400 images each.

\subsection{Texture Feature Extraction}
Texture features play a crucial role in characterizing the visual patterns and surface properties of objects in image analysis. These features are particularly important for wood chip MC prediction, where the texture of the chips correlates with moisture levels. In this study, we extracted five types of texture features: Haralick \citep{haralick1973textural}, First-Order Statistics (FOS), Fast Fourier Power Spectrum (FPS), Gray Level Run Length Matrix (GLRLM) \citep{galloway1974texture}, and Local Binary Patterns (LBP) \citep{ojala1996comparative}. Each of these methods captures different aspects of texture. Haralick features quantify spatial relationships between pixel intensities through co-occurrence matrices, while FOS captures basic statistical properties like mean and variance. FPS describes the frequency domain characteristics of texture, GLRLM assesses run lengths of similar intensities, and LBP focuses on the local intensity variations in a binary format. These texture descriptors collectively offer a comprehensive representation of the surface characteristics of wood chips, aiding in MC classification.
\subsubsection{Haralick Features}
Haralick texture features \citep{haralick1973textural} are a set of statistical features extracted from the gray-level co-occurrence matrix (GLCM), a matrix that quantifies how frequently pairs of pixel with specific values (gray levels) occur in a specified spatial relationship. GLCMs can be generated in multiple directions, and a rotation-invariant feature representation can be achieved by averaging the features computed from all directions. A detailed study on using Haralick features for wood chip texture analysis to predict MC was conducted by Rahman et al.~\cite{rahman2024moisture}. We adopted a similar feature extraction method and recommend that readers refer to Rahman et al.~\cite{rahman2024moisture} for details on Haralick feature extraction. However, the key distinction of our study compared to Rahman et al.~\cite{rahman2024moisture} is that we performed a comprehensive texture analysis using five different types of texture features across three sources of wood chips. Additionally, we combined these features to achieve higher accuracy in moisture prediction using the comprehensive set of machine-learning classifiers.
\subsubsection{FOS}
First-Order Statistics (FOS) are fundamental texture descriptors that focus on the statistical distribution of pixel intensities within an image. These features provide essential information regarding an image's overall brightness, contrast, and dispersion of intensity values without considering spatial relationships between pixels. By calculating basic statistical measures such as mean, variance, skewness, and entropy, FOS captures the global characteristics of the image, which are highly relevant for distinguishing different MC levels in wood chips. The following notations are used to compute FOS features, and their corresponding formula are provided in Table \ref{tab:fos}.

\begin{tabular}{ll}
     $I(i)$ & Intensity of pixel $i$\\
     $N$ & Total number of pixels\\
     $\mu$ & Mean of pixel intensities\\
     $\sigma$ & Standard deviation of pixel intensities\\
     $p(i)$ & Probability of occurrence of intensity $i$\\
     $L$ & Number of intensity levels in the image\\
     $\mathrm{Min}(I)$ & Minimum intensity value in the image\\
     $\mathrm{Max}(I)$ & Maximum intensity value in the image\\
     $\mathrm{Percentile}$ & The value below which a given percentage of pixel intensities falls\\
\end{tabular}

\begin{table}[h]
\centering
\caption{First Order Statistics (FOS) Features and Formulas}
\vspace{0.3cm}
\label{tab:fos}
\resizebox{0.8\linewidth}{!}{
\begin{tabular}{ll}
\hline
\textbf{Feature Name} & \textbf{Formula} \\ \hline
 Mean & $\mu = \frac{1}{N} \sum_{i=1}^{N} I(i)$ \\ 
 Variance & $\sigma^2 = \frac{1}{N} \sum_{i=1}^{N} (I(i) - \mu)^2$ \\  
 Median & $ \mathrm{Median}$ =  Middle value of the sorted pixel intensities \\  
 Mode & $ \mathrm{Mode}$ =  Most frequent pixel intensity \\  
 Skewness & $\frac{1}{N} \sum_{i=1}^{N} \left( \frac{I(i) - \mu}{\sigma} \right)^3$ \\  
 Kurtosis & $\frac{1}{N} \sum_{i=1}^{N} \left( \frac{I(i) - \mu}{\sigma} \right)^4$ \\  
 Energy & $\sum_{i=1}^{N} I(i)^2$ \\  
 Entropy & $-\sum_{i=1}^{L} p(i) \log(p(i))$ \\  
 MinimalGrayLevel & $ \mathrm{Min}(I)$ \\  
 MaximalGrayLevel & $ \mathrm{Max}(I)$ \\  
 CoefficientOfVariation & $\frac{\sigma}{\mu}$ \\  
 10Percentile &  10th percentile of pixel intensities \\  
 25Percentile &  25th percentile of pixel intensities  \\  
 75Percentile &  75th percentile of pixel intensities  \\  
 90Percentile &  90th percentile of pixel intensities  \\  
 HistogramWidth &  Max($I$) -  Min($I$) \\ \hline
\end{tabular}
}
\end{table}
\subsubsection{FPS}
The Fourier Power Spectrum (FPS) is a key tool for analyzing the frequency content of an image. It is derived from the Fourier Transform, which converts an image from the spatial domain (where pixel intensities are defined at specific positions) to the frequency domain (where information is represented in terms of its constituent frequencies). The following notations are used to compute the FPS features, and their corresponding formula is provided in Table \ref{tab:fps}.\\

\begin{tabular}{ll}
     $F(u, v)$ & Fourier coefficient at frequency $(u, v)$\\
     $I(x, y)$ & Intensity of the pixel at position $(x, y)$\\
     $M$ & Width of the image \\
     $N$ & Height of the image \\
     $u, v$ & Frequency coordinates \\
    $j$ & Imaginary unit ($j = \sqrt{-1}$) \\
     $\mathrm{Re}(F(u, v))$ & Real part of the Fourier coefficient \\
    $\mathrm{Im}(F(u, v))$ & Imaginary part of the Fourier coefficient \\
    $|F(u, v)|$ & Magnitude of the Fourier coefficient \\
    $P(u, v)$ & Power spectrum at frequency $(u, v)$ \\
\end{tabular}\\

\begin{equation}
F(u, v) = \sum_{x=0}^{M-1} \sum_{y=0}^{N-1} I(x, y) e^{-2\pi j \left( \frac{ux}{M} + \frac{vy}{N} \right)}
\end{equation}

\begin{equation}
|F(u, v)| = \sqrt{(\text{Re}(F(u, v)))^2 + (\text{Im}(F(u, v)))^2}
\end{equation}

\begin{equation}
P(u, v) = |F(u, v)|^2 = (\text{Re}(F(u, v)))^2 + (\text{Im}(F(u, v)))^2
\end{equation}

\begin{table}[h]
\centering
\caption{Fourier Power Spectrum (FPS) Features and Formulas}
\vspace{0.3cm}
\label{tab:fps}

\begin{tabular}{ll}

\hline
\textbf{Feature Name} & \textbf{Formula} \\ \hline
RadialSum & $P_r = \sum_{r} P(u, v)$ for each radial frequency bin $r$ \\ 
AngularSum & $P_\theta = \sum_{\theta} P(u, v)$ for each angular bin  $\theta$ \\ \hline
\end{tabular}
\end{table}

\subsubsection{GLRLM}
The Gray-Level Run-Length Matrix (GLRLM) \citep{galloway1974texture} is a statistical method used to describe the spatial arrangement of pixel intensities (gray levels) in an image. It quantifies the length of consecutive runs of pixels having the same gray level along specified directions (usually horizontal, vertical, or diagonal). These runs are important for understanding image texture, as they provide insight into the regularity, uniformity, and structure of an image. The following notations are used to compute the GLRLM features, and their corresponding formula are provided in Table \ref{tab:glrlm}.\\

\begin{tabular}{ll}
     $P(i, j)$& The value at gray level $i$ and run length $j$ in the GLRLM matrix \\
     $N_g$& Number of gray levels\\
     $N_r$& Number of run lengths\\
     $N_p$& Total number of pixels in the image\\
     $i$& Gray level index\\
     $j$& Run length index\\
\end{tabular}\\

\begin{table}[h]
\centering
\caption{GLRLM Features and Formulas}
\vspace{0.3cm}
\label{tab:glrlm}
\resizebox{0.7\linewidth}{!}{
\begin{tabular}{ll}
\hline
\textbf{Feature Name} & \textbf{Formula} \\ \hline
 ShortRunEmphasis & $\frac{1}{N_r} \sum_{i=1}^{N_g} \sum_{j=1}^{N_r} \frac{P(i, j)}{j^2}$ \\  
 LongRunEmphasis & $\frac{1}{N_r} \sum_{i=1}^{N_g} \sum_{j=1}^{N_r} j^2 P(i, j)$ \\  
 GrayLevelNonUniformity & $ \frac{1}{N_r} \sum_{i=1}^{N_g} \left( \sum_{j=1}^{N_r} P(i, j) \right)^2$ \\  
 RunLengthNonUniformity  & $ \frac{1}{N_r} \sum_{j=1}^{N_r} \left( \sum_{i=1}^{N_g} P(i, j) \right)^2$ \\  
 RunPercentage  & $ \frac{N_r}{N_p}$ \\  
 LowGrayLevelRunEmphasis  & $\frac{1}{N_r} \sum_{i=1}^{N_g} \sum_{j=1}^{N_r} \frac{P(i, j)}{i^2}$ \\  
 HighGrayLevelRunEmphasis  & $\frac{1}{N_r} \sum_{i=1}^{N_g} \sum_{j=1}^{N_r} i^2 P(i, j)$ \\  
 ShortRunLowGrayLevelEmphasis  & $\frac{1}{N_r} \sum_{i=1}^{N_g} \sum_{j=1}^{N_r} \frac{P(i, j)}{i^2 j^2}$ \\  
 ShortRunHighGrayLevelEmphasis & $\frac{1}{N_r} \sum_{i=1}^{N_g} \sum_{j=1}^{N_r} \frac{i^2 P(i, j)}{j^2}$ \\  
 LongRunLowGrayLevelEmphasis & $ \frac{1}{N_r} \sum_{i=1}^{N_g} \sum_{j=1}^{N_r} \frac{P(i, j) j^2}{i^2}$ \\  
 LongRunHighGrayLevelEmphasis  & $\frac{1}{N_r} \sum_{i=1}^{N_g} \sum_{j=1}^{N_r} i^2 j^2 P(i, j)$ \\ \hline
\end{tabular}
}
\end{table}

\subsubsection{LBP}
Local Binary Pattern (LBP) \citep{ojala1996comparative} is a texture descriptor used for analyzing local spatial patterns in an image. It is a simple but highly efficient method to characterize the texture by comparing each pixel with its surrounding neighbors. The basic idea behind LBP is to encode the local texture information by thresholding the neighborhood of each pixel relative to the center pixel and then converting this information into a binary number. The LBP histogram is then used to describe the image’s texture. The following notations are used to compute the LBP features, and their corresponding formula are provided in Table \ref{tab:lbp}.\\

\begin{tabular}{ll}
     $H_{R=a,P=b}(i)$& LBP histogram value at bin $i$ for radius $a$ and $b$ points\\
     $L$& Number of bins in the LBP histogram\\
\end{tabular}\\

\begin{table}[h]
\centering
\caption{LBP Features (Energy and Entropy) for Different Radii and Points}
\vspace{0.3cm}
\label{tab:lbp}
\begin{tabular}{ll}
\hline
\textbf{Feature Name} & \textbf{Formula} \\ \hline
 R\_1\_P\_8\_Energy & $\sum_{i=1}^{L} H_{R=1,P=8}(i)^2$ \\  
 R\_1\_P\_8\_Entropy & $ - \sum_{i=1}^{L} H_{R=1,P=8}(i) \log(H_{R=1,P=8}(i))$ \\  
 R\_2\_P\_16\_Energy & $ \sum_{i=1}^{L} H_{R=2,P=16}(i)^2$ \\  
 R\_2\_P\_16\_Entropy & $ - \sum_{i=1}^{L} H_{R=2,P=16}(i) \log(H_{R=2,P=16}(i))$ \\  
 R\_3\_P\_24\_Energy & $\sum_{i=1}^{L} H_{R=3,P=24}(i)^2$ \\  
 R\_3\_P\_24\_Entropy & $- \sum_{i=1}^{L} H_{R=3,P=24}(i) \log(H_{R=3,P=24}(i))$ \\  \hline
\end{tabular}

\end{table}

\subsection{Predictive Models \& Evaluation Metrics}
Machine learning (ML) predictive models have shown great promise across various domains, delivering strong prediction results using manually extracted features. In this study, we utilized a total of twelve machine learning (ML) models, comprising six classical ML models, five ensemble models, and a voting classifier \citep{kumar2017prediction}. In previous studies, these models have proven effective in classifying wood chip MC, yielding promising results \citep{daassi2017wood, suthar2021multiclass, daassi2018moisture, tiitta2020classification}. The reason for selecting a diverse set of models is to conduct a comprehensive comparison and determine the best-performing model for predicting wood chip MC. The list of classical ML models is comprised of K-Nearest Neighbors (KNN) \citep{peterson2009k}, Logistic Regression (LR) \citep{kleinbaum2002logistic}, Naive Bayes (NB) \citep{zhang2004optimality}, Decision Tree (DT) \citep{song2015decision}, Support  Vector Machine (SVM) \citep{steinwart2008support}, and Neural Network (NN) \citep{taud2018multilayer}. The list of ensemble models included Random Forest (RF) \citep{breiman2001random}, Bagging \citep{breiman1996bagging}, Extra Tree (ET) \citep{ossai2022glcm}, AdaBoost \citep{ying2013advance}, and Gradient Boosting \citep{natekin2013gradient}. 

We approached the MC prediction as a multi-class classification task, categorizing the data into three MC levels: \textit{dry}, \textit{medium}, and \textit{wet}, following the work of Rahman et al.~\cite{rahman2025moistnet}. To evaluate model performance, accuracy, precision, recall, and F1-score were selected as the primary metrics. These metrics were calculated using Equations (\ref{eqn:acc}) - (\ref{eqn:f1}), where TP, TN, FP, and FN correspond to true positive, true negative, false positive, and false negative values, respectively.
\begin{equation}
\label{eqn:acc}
    Accuracy = \frac{TP + TN}{TP + TN + FP +FN}
\end{equation}
\begin{equation}
\label{eqn:pre}
    Precision = \frac{TP}{TP + FP}
\end{equation}
\begin{equation}
\label{eqn:rec}
    Recall = \frac{TP}{TP + FN}
\end{equation}
\begin{equation}
\label{eqn:f1}
    F1\: score = \frac{2\times Precision \times Recall}{Precision + Recall}
\end{equation}

\subsection{Proposed AdaptMoist Model}

\begin{figure}[h]
    \centering
    \includegraphics[width=\linewidth]{ 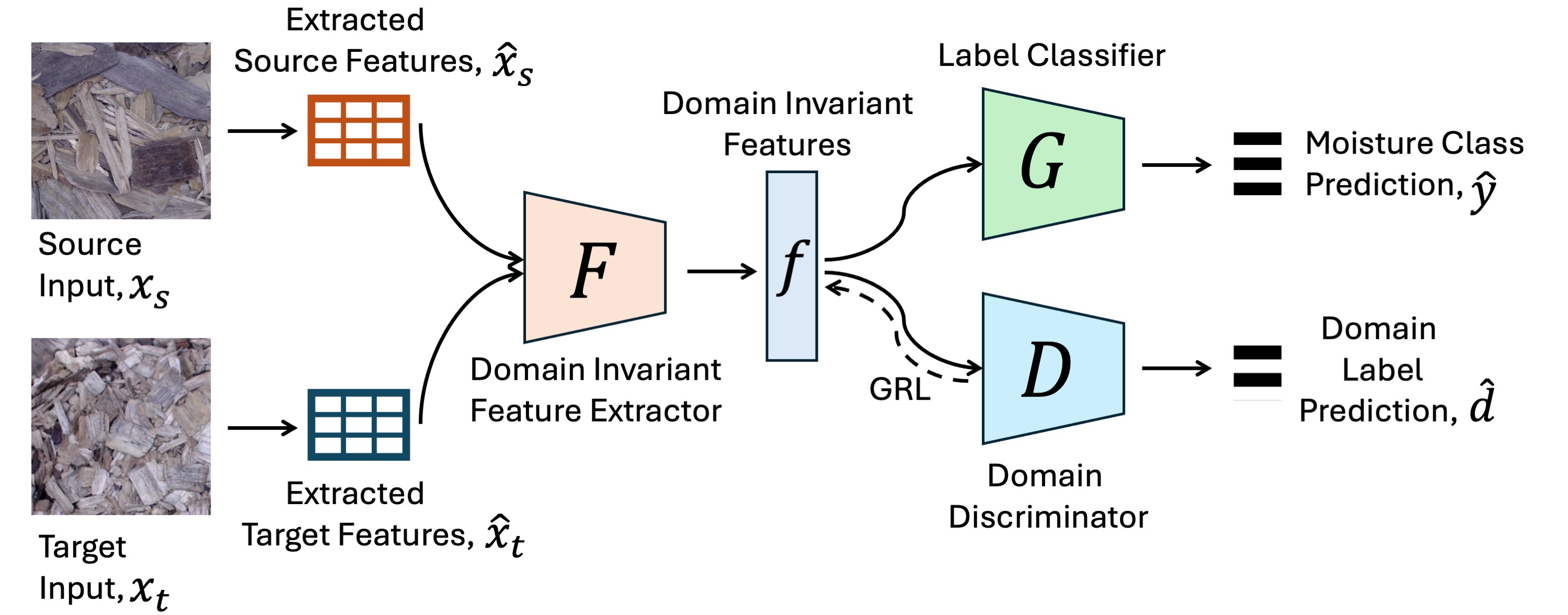}
    \caption{Proposed AdaptMoist method to integrate unsupervised domain adaptation for predicting wood chip MC classes. Texture features $\hat{\mathbf{x}}_s$ and $\hat{\mathbf{x}}_t$ are initially extracted from the source and target raw images $\mathbf{x}_s$ and $\mathbf{x}_t$, respectively. These features serve as the input to a feature extractor network $F$, which produces the domain invariant features $\mathbf{f}$. A label classifier network $G$ and a domain discriminator network $D$ then take $\mathbf{f}$ as input and predict the moisture class $\hat{y}$ and $\hat{d}$, respectively. This framework learns domain-invariant feature $\mathbf{f}$ with the help of a gradient reversal layer in the domain discriminator $D$.} 
    \label{fig:process}
\end{figure}

Our proposed AdaptMoist model for wood chip MC prediction is based on the Domain-Adversarial Neural Network (DANN) \citep{ganin2016domain} framework. The goal is to effectively transfer knowledge from a labeled source domain to an unlabeled target domain by learning domain-invariant features that are robust to domain shifts. As shown in Figure \ref{fig:process}, the architecture consists of three key components: a feature extractor (\(F\)), a label classifier (\(G\)), and a domain discriminator (\(D\)). The feature extractor learns a shared representation from both source and target domains, while the discriminator plays an adversarial role in ensuring domain invariance.

The feature extractor (\(F\)) processes both source input $(\mathbf{x}_s)$ and target input $(\mathbf{x}_t)$ to learn domain-invariant features, denoted as $\mathbf{f}$. However, instead of using raw image input, the AdaptMoist model uses the extracted texture features $(\mathbf{\hat{x}}_s)$ and $(\mathbf{\hat{x}}_t)$. The domain invariant or shared domain features are then fed into two branches: the label classifier (\(G\)) and the domain discriminator (\(D\)). The goal of this component is to ensure that the learned features are rich enough to predict MC while being agnostic to the domain from which the data originates.

The label classifier (\(G\)) takes the domain-invariant features \(\mathbf{f}\) from the feature extractor and predicts the MC class for the source domain. The prediction \(\hat{y}\) represents the classification of MC as \textit{dry}, \textit{medium}, or \textit{wet}. The loss for the label classifier is the standard categorical cross-entropy:
\begin{equation}
\mathcal{L}_{\text{label}} = - \frac{1}{N_s} \sum_{i=1}^{N_s} \sum_{k=1}^{C} y_{ik} \log(\hat{y}_{ik}),
\end{equation}

where \(N_s\) is the number of samples in the source domain, \(C\) is the number of classes, \(y_{ik}\) is the true label of the \(i\)-th sample for class \(k\), and \(\hat{y}_{ik}\) is the predicted probability of class \(k\).

The domain discriminator (\(D\)) is responsible for distinguishing whether the feature \(\mathbf{f}\) originates from the source domain or the target domain. To ensure that the learned features are domain-invariant, we employ a Gradient Reversal Layer (GRL) between the feature extractor and the domain discriminator. The GRL inverts the gradients during backpropagation, encouraging the feature extractor to learn features that make it difficult for the domain discriminator to classify the domain correctly.

The domain label prediction \(\hat{d}\) is generated by the discriminator, where a domain label of 0 represents the source domain and a label of 1 represents the target domain. The loss for the domain discriminator is defined as:

\begin{equation}
\mathcal{L}_{\text{domain}} = - \frac{1}{N_s + N_t} \sum_{i=1}^{N_s + N_t} \left[ d_i \log(\hat{d}_i) + (1 - d_i) \log(1 - \hat{d}_i) \right],
\end{equation}

where \(N_s\) and \(N_t\) represent the number of samples in the source and target domains, respectively, and \(d_i\) is the true domain label of the \(i\)-th sample.

The total loss function for the proposed domain adaptation model combines the label classification loss and the domain discrimination loss. The two losses are balanced by a weight parameter \(\lambda\), which controls the trade-off between domain adaptation and classification accuracy. The total loss is expressed as:

\begin{equation}
\mathcal{L}_{\text{total}} = \mathcal{L}_{\text{label}} + \lambda \mathcal{L}_{\text{domain}}.
\end{equation}

By optimizing this total loss, the AdaptMoist model learns features that are both discriminative for MC classification and invariant to the domain, ensuring robust performance across different wood chip sources.

In unsupervised domain adaptation (UDA) model training, we do not have access to the target domain labels. Therefore, model validation and finding out the best weight for the trained model that would perform the best on the target domain is challenging. One way to do this is to use a portion of the source domain for validation, which could be misleading due to the distribution shift in the source and target domain. Another way is to use a portion of the target domain labels to validate the model, which again violates the unsupervised setting principle. Therefore, several unsupervised ways or validators have been proposed for UDA \citep{musgrave2022three}. Adjusted Mutual Information (AMI) has shown potential as a validator in Musgrave et al.~\cite{musgrave2022three}.

In the AdaptMoist model, we employed the Adjusted Mutual Information (AMI) as a model-saving criterion to monitor clustering alignment between the source and target domains during training. AMI is particularly useful for assessing the similarity between two label assignments, taking into account the chance of random labeling. Unlike traditional clustering metrics, AMI is adjusted for the agreement expected by chance. This makes it a robust metric in domain adaptation scenarios where the labels in the target domain are unknown.

Given two sets of labels, \(\mathcal{U} = \{u_1, u_2, \dots, u_k\}\) and \(\mathcal{V} = \{v_1, v_2, \dots, v_k\}\), representing the predicted labels from the AdaptMoist model and pseudo-labels from the KMeans clustering method, respectively, the AMI is defined as follows:

\begin{equation}
    \text{AMI}(\mathcal{U}, \mathcal{V}) = \frac{\text{MI}(\mathcal{U}, \mathcal{V}) - \mathbb{E}[\text{MI}(\mathcal{U}, \mathcal{V})]}{\max(H(\mathcal{U}), H(\mathcal{V})) - \mathbb{E}[\text{MI}(\mathcal{U}, \mathcal{V})]},
\end{equation}

where \(\text{MI}(\mathcal{U}, \mathcal{V})\) is the mutual information between \(\mathcal{U}\) and \(\mathcal{V}\), \(H(\mathcal{U})\) and \(H(\mathcal{V})\) are the entropies of \(\mathcal{U}\) and \(\mathcal{V}\), and \(\mathbb{E}[\text{MI}(\mathcal{U}, \mathcal{V})]\) is the expected mutual information, which accounts for the chance alignment between the clusters.

The mutual information \(\text{MI}(\mathcal{U}, \mathcal{V})\) is calculated as:

\begin{equation}
    \text{MI}(\mathcal{U}, \mathcal{V}) = \sum_{i=1}^{|\mathcal{U}|} \sum_{j=1}^{|\mathcal{V}|} P(u_i, v_j) \log \frac{P(u_i, v_j)}{P(u_i)P(v_j)},
\end{equation}

where \(P(u_i, v_j)\) is the joint probability of \(u_i\) and \(v_j\), and \(P(u_i)\) and \(P(v_j)\) are the marginal probabilities of \(u_i\) and \(v_j\), respectively.

The AMI score ranges from 0 to 1, where 1 indicates perfect alignment between the labels, and 0 indicates no more agreement than would be expected by random chance. Therefore, the higher the AMI score, the better. During training, we implemented a custom callback in TensorFlow to compute the AMI score at each epoch, comparing the predicted target domain labels to the pseudo-labels generated for the target domain using the KMeans clustering algorithm. We also used Fuzzy CMeans \citep{bezdek2013pattern} to generate the pseudo labels, however, no significant changes have been observed. That is why we chose the KMeans clustering algorithm for its simplicity and effectiveness. This metric allowed us to monitor the degree of alignment between the source and target domains as the model trained, enabling us to track the model's adaptation performance during training.

\subsection{Baselines for Domain Adaptation}
%DANN, ADDA, Deep CORAL, MDD, WDGRL, CCSA

In this study, we compared our proposed model against several well-established domain adaptation baseline models. Since the texture data is of a tabular format, we considered models with less complexity, avoiding recently published transformer-based works. These baselines represent a diverse set of approaches to domain adaptation, ranging from adversarial learning (DANN \citep{ganin2016domain}, ADDA \citep{tzeng2017adversarial}, WDGRL \citep{shen2018wasserstein}) and statistical alignment (Deep CORAL \citep{sun2016deep}) to margin-based learning (MDD \citep{zhang2019bridging}) and contrastive alignment (CCSA \citep{motiian2017unified}). The following provides a brief overview of the baseline models used:

\textbf{Source Only}: This baseline involves training a model solely on the labeled source domain data without utilizing any domain adaptation techniques. The model is evaluated directly on the target domain, serving as a lower bound to measure the effectiveness of domain adaptation methods.

\textbf{Domain-Adversarial Neural Network (DANN)} \citep{ganin2016domain}: DANN is a widely-used domain adaptation method that introduces a domain classifier and a Gradient Reversal Layer (GRL) to align feature representations from the source and target domains. The GRL reverses the gradients from the domain classifier to ensure that the feature extractor learns domain-invariant features. DANN optimizes the model through joint learning of classification and domain discrimination tasks.

\textbf{Adversarial Discriminative Domain Adaptation (ADDA)} \citep{tzeng2017adversarial}: ADDA leverages adversarial learning to align feature distributions between the source and target domains. Unlike DANN, ADDA separates the learning of feature extractors for the source and target domains, with a discriminator trained to differentiate between the two. ADDA utilizes adversarial training to encourage the target feature space to align with the source feature space while keeping classifiers independent.

\textbf{Deep CORAL (Deep Correlation Alignment)} \citep{sun2016deep}: Deep CORAL minimizes domain shift by aligning the second-order statistics of the source and target feature distributions. Specifically, it computes the covariance of the source and target features and penalizes their difference using the Frobenius norm. This method aims to reduce the domain discrepancy without requiring labeled target data.

\textbf{Margin Disparity Discrepancy (MDD)} \citep{zhang2019bridging}: MDD addresses domain adaptation by explicitly minimizing the disparity between source and target classification margins. This method optimizes a margin disparity loss, ensuring that the classifier generalizes well to the target domain. MDD emphasizes the confidence of predictions for both domains, leading to better performance on target domain samples.

\textbf{Wasserstein Distance Guided Representation Learning (WDGRL)} \citep{shen2018wasserstein}: WDGRL leverages the Wasserstein distance to align the source and target distributions. It minimizes the Wasserstein distance between source and target feature representations to reduce domain discrepancy. Additionally, a domain critic is trained to compute the Wasserstein distance while the feature extractor is updated to minimize it.

\textbf{Classification and Contrastive Semantic Alignment (CCSA)} \citep{motiian2017unified}: CCSA aims to learn domain-invariant features by using a contrastive loss that minimizes the distance between similar source and target domain samples while maximizing the distance between dissimilar ones. This method leverages paired samples from the source and target domains, encouraging the model to learn a common feature space that discriminates between different classes regardless of domain.

\subsection{Implementation Details}
In the first set of experiments, we evaluated a list of twelve machine learning models for the task of predicting wood chip MC levels, categorized into \textit{dry}, \textit{medium}, and \textit{wet}. The experiments were conducted using texture features extracted from the raw wood chip images. The features were standardized to ensure zero mean and unit variance. We utilized a stratified K-fold cross-validation approach with four folds to maintain the balance of MC categories across both the training and validation sets. Each model was trained and validated using these splits, and the results were averaged across the folds to ensure robust performance evaluation. Rahman et al.~\cite{rahman2025moistnet} also used a 4-fold cross-validation strategy. All the models were used with similar settings as used by Rahman et al.~\cite{rahman2024moisture}. However, in this study, in the Voting Classifier, the Logistic regression, SVM, and the Multi-Layer Perceptron network are used as voters. The Voting Classifier used a soft voting strategy, which refers to a technique where the predicted probabilities (instead of the predicted labels) of each classifier are averaged, and the class label with the highest average probability is chosen as the final prediction. Soft voting often works better than hard voting (where each classifier votes for a class, and the majority wins) because it takes the confidence of each classifier into account, allowing more balanced decision-making. The implementation code has been made public for reproducibility\footnote{\url{https://github.com/abdurrahman1828/AdaptMoist}}.

The baselines for the domain adaptation models were implemented using the Adapt library\footnote{\url{https://adapt-python.github.io/adapt/index.html}}. For the source-only training, we employed the best-performing model derived from the machine learning models, which is the voting classifier. The AdaptMoist model consists of three component networks: (1) Feature Extractor $F$: A straightforward feed-forward network that converts the input texture features into a 32-dimensional representation using a single dense layer; (2) Label Classifier $G$: Another feed-forward network that includes a hidden layer with 16 units (ReLU activation) followed by an output layer with 3 units and softmax activation for classification; and (3) Domain Discriminator $D$: A feed-forward network comprised of a hidden layer with 16 units (ReLU activation) followed by a single output unit with sigmoid activation. 

We introduced a custom callback to compute the Adjusted Mutual Information (AMI) score at the end of each epoch. The pseudo-labels used for the AMI calculation were generated through KMeans clustering on the encoded target features. The number of clusters was set to three, matching the number of classes in the dataset. The value of \(\lambda\) was configured to 0.5 in the AdaptMoist model to balance the loss components. We trained the model for 30 epochs with a batch size of 2 and utilized the AMI callback. The AMI callback's initial warm-up iterations were set to 15. We saved the best model based on the AMI callback and used it to evaluate the predictive performance of the target data. 

\section{Results and Discussions}
In this section, we presented the results from the traditional machine learning model using individual data without domain adaptation alongside those obtained from the AdaptMoist domain adaptation model. We also examined various aspects of the results and their interpretations. 

\subsection{Performance of Machine Learning Methods on Individual Data}
Initially, the individual datasets were utilized to train and evaluate the traditional machine-learning models. These tests also include the results based on the feature extraction methods: Haralick, FOS, FPS, GLRLM, LBP, and their combination. The outcomes are displayed in thirty tables, ranging from Table \ref{tab:s1_haralick} to Table \ref{tab:s2b2_combined}. However, the best performance is summarized in Table \ref{tab:s_sum}, which outlines the sources and the top-performing classifier among them. We reported classification performance through accuracy, precision, recall, and F1-score in Table \ref{tab:s_sum}. From S1, the voting classifier achieved the highest accuracy of 0.95, surpassing the previous benchmark accuracy of 0.91 established by Rahman et al.~\cite{rahman2025moistnet} and 0.88 as obtained by Rahman et al.~\cite{rahman2024moisture}. 
\begin{table}[h]
\centering
\caption{Performance metrics for different datasets and the best-performing classifiers.}
\label{tab:s_sum}
\vspace{0.3cm}
\resizebox{\linewidth}{!}{
\begin{tabular}{cccccc}
\hline
\textbf{Dataset} & \textbf{Best Classifier} & \textbf{Accuracy} & \textbf{Precision} & \textbf{Recall} & \textbf{F1-score} \\ \hline
S1    & Voting Classifier & 0.9525 (0.0119) & 0.9531 (0.0116) & 0.9525 (0.0119) & 0.9525 (0.0119) \\ 
S2    & Neural Network    & 0.9175 (0.0328) & 0.9185 (0.0324) & 0.9175 (0.0328) & 0.9177 (0.0326) \\ 
S3    & Neural Network    & 0.9500 (0.0183) & 0.9509 (0.0174) & 0.9500 (0.0183) & 0.9497 (0.0186) \\ 
S2-B1 & Logistic Regression & 0.9300 (0.0294) & 0.9313 (0.0282) & 0.9300 (0.0294) & 0.9301 (0.0291) \\ 
S2-B2 & Voting Classifier  & 0.9300 (0.0163) & 0.9320 (0.0155) & 0.9300 (0.0163) & 0.9300 (0.0161) \\ \hline
\end{tabular}
}
\end{table}
\begin{figure}[h]
    \centering
    \includegraphics[width=1\linewidth]{ 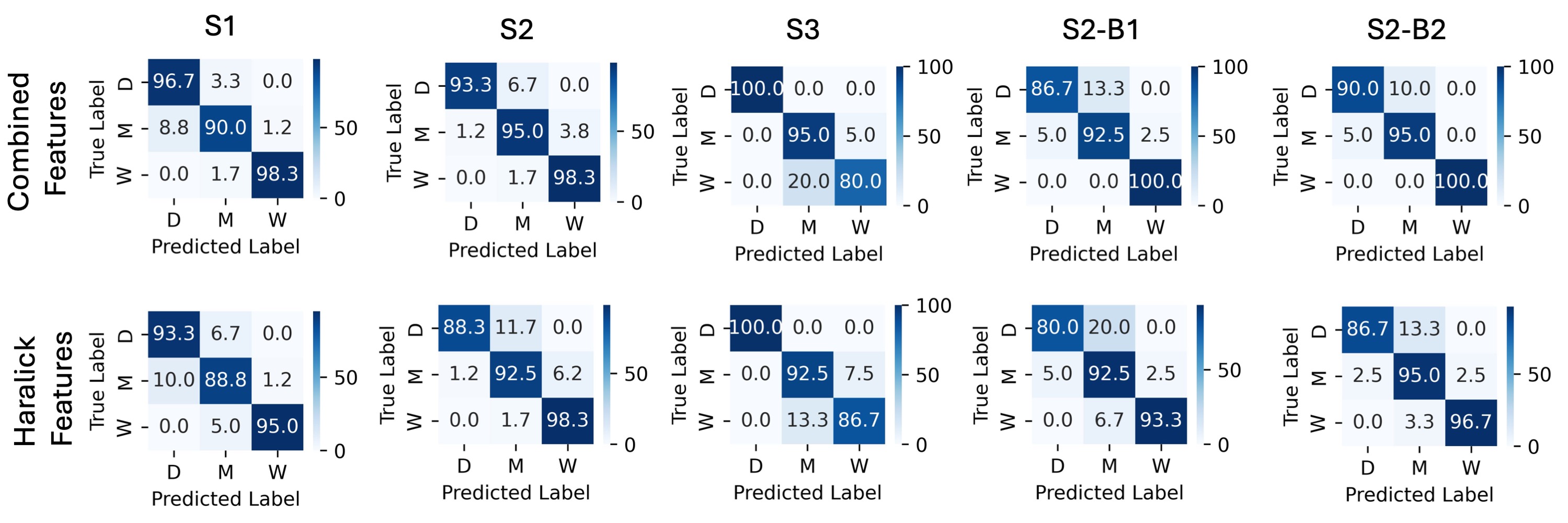}
    \caption{Confusion matrices of Haralick and combined features with the voting classifier on datasets from different sources and batches. Since we conducted k-fold cross-validation, confusion matrices for only the first folds have been illustrated. D, M, and W indicate dry, medium, and wet classes, respectively.}
    \label{fig:conf_features}
\end{figure}

\begin{table}[h]
\centering
\caption{F1-Score for different datasets using various texture feature extraction methods with mean and standard deviation (in brackets).}
\label{fig:feature-wise}
\vspace{0.3cm}
\resizebox{\linewidth}{!}{
\begin{tabular}{ccccccc}
\hline
\textbf{Dataset} & \textbf{Haralick} & \textbf{FOS} & \textbf{FPS} & \textbf{GLRLM} & \textbf{LBP} & \textbf{Combined} \\ \hline
S1    & 0.9224 (0.0278) & 0.7750 (0.0310) & 0.5533 (0.0177) & 0.8086 (0.0245) & 0.7749 (0.0511) & 0.9525 (0.0119) \\ 
S2    & 0.9151 (0.0173) & 0.7995 (0.0117) & 0.7230 (0.0276) & 0.8417 (0.0261) & 0.7389 (0.0309) & 0.9177 (0.0326) \\ 
S3    & 0.9498 (0.0143) & 0.8753 (0.0373) & 0.8034 (0.0391) & 0.8352 (0.0252) & 0.8145 (0.0300) & 0.9497 (0.0186) \\ 
S2-B1 & 0.9201 (0.0260) & 0.8141 (0.0246) & 0.7280 (0.0211) & 0.7883 (0.0337) & 0.7705 (0.0212) & 0.9301 (0.0291) \\ 
S2-B2 & 0.9275 (0.0097) & 0.8037 (0.0146) & 0.7123 (0.0182) & 0.8212 (0.0489) & 0.7564 (0.0421) & 0.9300 (0.0161) \\ \hline
\end{tabular}
}

\end{table}

We used the same number of folds in the k-fold cross-validation with the same random seed to split the dataset, making the results comparable to those of Rahman et al.~\cite{rahman2025moistnet} and Rahman et al.~\cite{rahman2024moisture}. Similarly, an MLP network achieved the highest accuracy of 0.92 in S2, surpassing the benchmarks of 0.90, as shown in Rahman et al.~\cite{rahman2025moistnet}, and 0.89, as shown in Rahman et al.~\cite{rahman2024moisture}. Furthermore, in S2-B1 and S2-B2, this study outperformed MoistNet~\citep{rahman2025moistnet} by 3.33\% and 10.71\%, respectively. These results are also supported by the confusion matrices shown in Figure \ref{fig:conf_features}, which illustrate the confusion matrices for both combined and Haralick features across all the datasets. We observed a pattern of higher misclassification in the dry vs. medium pair in the S1 and S2 datasets, similar to the results shown in the confusion matrices of MoistNet~\citep{rahman2025moistnet}. However, in S3, the most misclassifications occurred in the medium vs. wet pair. Since Rahman et al.~\cite{rahman2025moistnet} did not run tests on the S3 dataset, we could not compare results for this data.

We also reported the summary of the performance of the best-performing machine learning models according to feature type in Table \ref{fig:feature-wise}. The Haralick features yielded the highest accuracy among the five extracted feature sets, while the FPS features produced the lowest accuracy. However, when all features were combined, the resulting feature set achieved the best overall accuracy. This result suggests that, although some individual feature sets may not achieve optimal performance independently, their combination effectively enhances model accuracy, indicating complementary information across feature types.

\subsection{Performance of AdaptMoist in Wood Chip Domain Adaptation}
To evaluate the domain adaptation performance of the proposed AdaptMoist and the benchmark methods, we considered the images collected from each source as a domain. We trained each of these domain adaptation methods on one domain (the source domain) and tested them on another domain (the target domain) to assess the generalization ability of the models. Importantly, these tests were performed in an unsupervised domain adaptation setting, meaning the target domain labels were not used during training. The results of these experiments are presented in Table \ref{tab:da_results}. With three distinct wood chip sources, we have three domains and six domain adaptation tasks. Additionally, we included two batches from one source as separate domains to assess the model’s generalizability across different batches from the same source. 

\begin{table}[h]
\centering
\caption{Accuracy of domain adaptation models on wood chip datasets from different sources and batches. S1$\to$S2 indicates training the models on S1 and testing on S2.}
\vspace{0.3cm}
\label{tab:da_results}
\resizebox{\linewidth}{!}{
\begin{tabular}{l|cccccc|cc|c}
\hline
Models                                                        & S1$\to$S2 & S1$\to$S3 & S2$\to$S1 & S2$\to$S3 & S3$\to$S1 & S3$\to$S2 & S2B1$\to$S2B2 & S1B2$\to$S2B1 & Avg.\\ \hline
Source Only & 0.54 & 0.72 & 0.55 & 0.30  & 0.32 & 0.35 & 0.89 & 0.85 & 0.57 \\
DANN  \citep{ganin2016domain}                                                 &  0.71 & 0.66 & 0.70 & 0.65 & 0.72 & 0.72 & \textbf{0.90} & \textbf{0.86} & 0.74 \\
ADDA     \cite{tzeng2017adversarial}                                             & 0.60  &  0.62 & 0.51 & 0.57 & 0.65 & 0.69 & 0.88 & 0.82  & 0.67\\
Deep CORAL \citep{sun2016deep} & 0.58 & 0.52 & 0.61 & 0.48 & 0.40 & 0.48 & 0.61 & 0.60 & 0.54\\
MDD   \citep{zhang2019bridging}                                                & 0.62 & 0.66 & 0.57  & 0.48 & 0.50 & 0.42 & 0.88 & 0.83 & 0.62 \\
WDGRL  \citep{shen2018wasserstein}                                               &  0.74 & 0.69 & 0.62 & 0.59 & 0.59 & 0.73 & 0.89 & 0.82 & 0.71\\
CCSA   \citep{motiian2017unified}                                                & 0.79 & 0.82 & 0.68 & 0.72 & 0.72 & 0.77 & 0.89 & 0.85 & 0.78 \\\hline
Proposed AdaptMoist &    \textbf{0.80}       &     \textbf{0.86}      &     \textbf{0.72}      &    \textbf{0.76}       &    \textbf{0.74}       &   \textbf{0.78}        &    0.87           &      0.84        & \textbf{0.80} \\ \hline
\end{tabular}
}
\end{table}

The `Source Only' model refers to training solely on the source domain without any domain adaptation. As shown in Table \ref{tab:da_results}, the performance of the model without domain adaptation is significantly low, with an average accuracy of 0.57 across all domain adaptation tasks. Among the domain adaptation models, AdaptMoist achieved the highest accuracy, with an average of 0.80 across tasks. The CCSA \citep{motiian2017unified} model followed closely, attaining the second-best result with an average accuracy of 0.78, which is 2.5\% lower than that of the proposed AdaptMoist model. The accuracy of the domain adaptation task across the batches from Source 2 is notably higher than that across different sources. This is because the batches of chips come from the same source but from different locations within the woodchip pile, leading to a smaller shift in the data distribution between the two batches. In contrast, domain adaptation tasks across different sources are more challenging due to a greater shift in the data distribution. DANN \citep{ganin2016domain} demonstrated the highest performance for the domain adaptation task across the batches from Source 2. This suggests that DANN is more effective when the distribution shift is smaller. Conversely, it is clear that our proposed AdaptMoist model performs effectively in tasks across different sources where the distribution shift is larger.

\begin{figure}[h]
    \centering
    \includegraphics[width=1\linewidth]{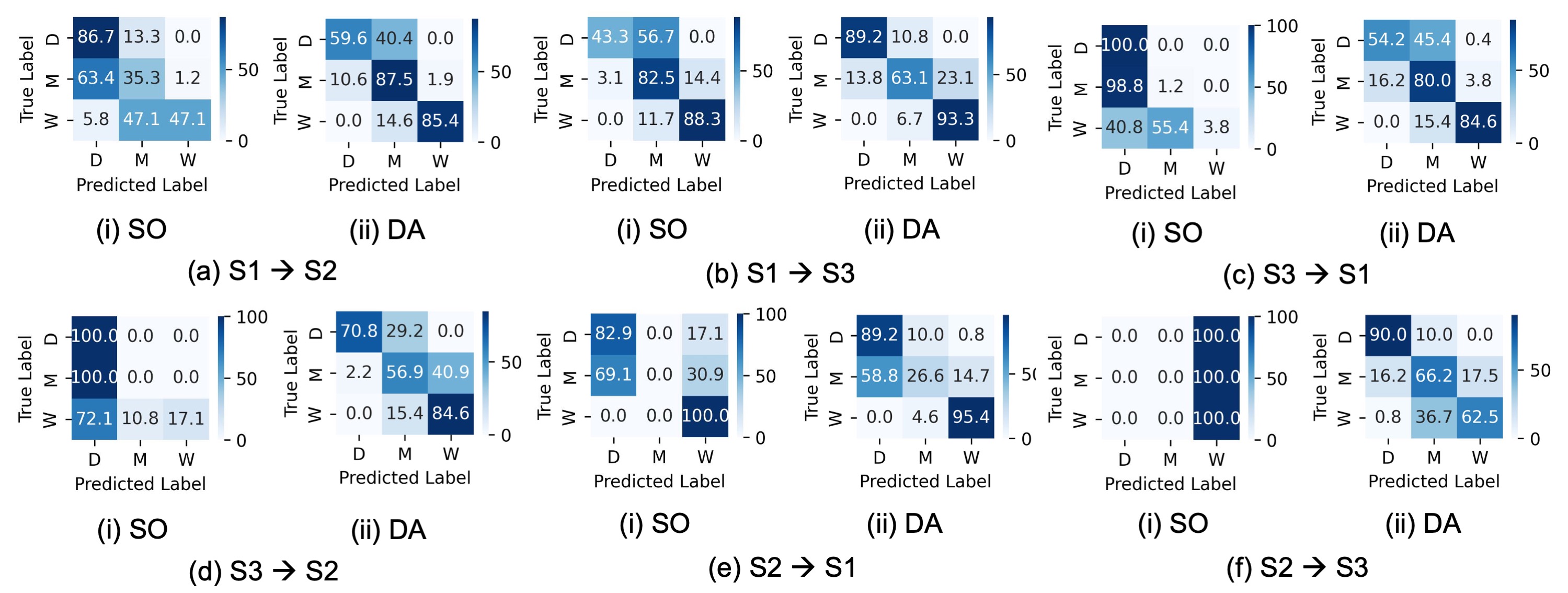}
    \caption{Confusion matrices for different domain adaptation scenarios. SO and DA indicate source-only training (without domain adaptation) and domain adaptation, respectively. S1 $\to$ S2 means trained on S1 (source) and tested on S2 (target).}
    \label{fig:conf_features}
\end{figure}
Figure \ref{fig:conf_features} presents the confusion matrices for the source-only training (denoted as SO) and the proposed AdaptMoist model (denoted as DA) across the six domain adaptation tasks. It is clear that the source-only model struggles to accurately classify the samples, often predicting the majority of samples as a single class (e.g., in the S3$\to$S2 and S2$\to$S3 tasks). In contrast, AdaptMoist shows a significant improvement in domain adaptation performance, particularly when predicting classes for new sources, as demonstrated in Figure \ref{fig:conf_features}.

Up to this point, the domain adaptation experiments initially focused on the Haralick texture features. We also conducted additional experiments to explore how other texture features might contribute to generalization. Table \ref{tab:da_models} presents the performance of the proposed AdaptMoist on the wood chip datasets using various texture features. Surprisingly, only the Haralick features demonstrated consistent performance, while other features struggled to achieve effective domain adaptation. This may be due to the absence of transferable (shared) features in textures such as FOS, FPS, GLRLM, and LBP. Even the combined feature set did not significantly improve domain adaptation across different sources, likely due to the inclusion of these less transferable features. However, for across-batch domain adaptation, the combined features showed positive results, which could be attributed to the smaller distribution shift between batches.
\begin{comment}

\begin{table}[h]
\centering
\caption{Results on domain adaptation. SO and DA refer to source only and domain adaptation training protocols. }
\vspace{0.3cm}
\label{tab:da}
\begin{tabular}{ccccccccc}
\hline
Metric & \multicolumn{2}{c}{Accuracy} & \multicolumn{2}{c}{Precision} & \multicolumn{2}{c}{Recall} & \multicolumn{2}{c}{F1-score} \\ \hline
Training Protocol    & SO   & DA   & SO   & DA   & SO   & DA   & SO   & DA   \\ \hline
S1$\to$S2 & 0.54 & 0.79 & 0.61 & 0.81 & 0.54 & 0.79 & 0.53 & 0.78 \\
S2$\to$S1 & 0.55 & 0.66 & 0.33 & 0.69 & 0.55 & 0.66 & 0.41 & 0.62 \\
S1$\to$S3 & 0.72 & 0.80 & 0.77 & 0.81 & 0.72 & 0.80 & 0.71 & 0.79 \\
S3$\to$S1 & 0.32 & 0.74 & 0.42 & 0.75 & 0.32 & 0.74 & 0.19 & 0.74 \\
S2$\to$S3 & 0.30 & 0.72 & 0.09 & 0.72 & 0.30 & 0.72 & 0.14 & 0.72 \\
S3$\to$S2 & 0.35 & 0.69 & 0.40 & 0.72 & 0.35 & 0.69 & 0.24 & 0.70 \\ \hline
S2-B1$\to$S2-B2 & 0.89 & 0.87 & 0.89 & 0.87 & 0.89 & 0.87 & 0.89 & 0.87 \\
S2-B2$\to$S2-B1 & 0.85 & 0.84 & 0.86 & 0.85 & 0.85 & 0.84 & 0.85 & 0.84 \\ \hline
\end{tabular}
\end{table}

\end{comment}

\subsection{Discussion}
%Summary of key findings
Wood chips are dense, small objects whose color and texture can change with varying moisture levels. In this study, we thoroughly evaluated the effectiveness of five different texture features and their combinations for predicting wood chip moisture levels using a set of machine learning classifiers. The results highlight the promising potential of this approach for real-time moisture prediction in industrial applications. While Rahman et al.~\cite{rahman2024moisture} demonstrated the potential of texture features for wood chip MC prediction with an accuracy of 0.88, we achieved an improved accuracy of 0.95 by combining five distinct texture features. This represents the state-of-the-art result of this wood chip dataset. Upon analyzing the confusion matrices, we observed that distinguishing between dry and medium samples is more challenging than differentiating between medium and wet samples. This finding aligns with the observations of Rahman et al.~\cite{rahman2025moistnet, rahman2024moisture}. To address this challenge, potential solutions could involve incorporating more samples from this moisture range, employing data augmentation techniques to increase the sample size, and assigning greater weight to these samples during training.

The heterogeneity of wood chips arises from various factors, including shape, color, plant type, cutting method, origin, time of production, and more~\citep{rahman2024comprehensive}. These variations pose significant challenges for data-driven indirect approaches, which often struggle to generalize across all types of wood chips. The problem becomes particularly challenging when there is a substantial difference between the data used for model calibration (training) and the unseen incoming wood chip samples. Traditional machine learning methods, including deep learning approaches, often fail to generalize effectively in such cases. This is where our proposed domain adaptation approach, AdaptMoist, proves beneficial. The AdaptMoist framework employs an adversarial training method to extract transferable features from the wood chip data while simultaneously learning to differentiate between moisture level classes. Additionally, the proposed AMI callback effectively halts the training, preventing overfitting to the source data and thereby improving domain adaptation performance. This enables accurate prediction of moisture classes for unseen wood chip samples.

Although combined texture features demonstrated superior performance on data from individual sources, they faced significant challenges when applied in domain adaptation, struggling to effectively transfer knowledge. Other features, such as FOS, FPS, GLRLM, and LBP, underperformed in the domain adaptation tasks. The most promising results, however, were achieved using Haralick features, suggesting a higher degree of transferability. Future studies could focus on identifying the specific Haralick features that exhibit better transferability and those that are more discriminative in terms of class differentiation. 

The impact of our proposed solution is substantial, given the rapidly growing global wood chip market and the increasing number of industries reliant on wood chip-based products. In real-world applications, data heterogeneity presents a major challenge, and for materials like wood chips, this can often lead to the failure of traditional data-driven approaches. Our proposed AdaptMoist framework, if deployed on mobile devices (e.g., smartphones), offers a transformative solution for large-scale industries to measure MC efficiently and accurately. The conventional oven-drying method, while reliable, takes hours to yield moisture content results, making it impractical for real-time process control. To address this, industries utilize moisture analyzers that operate on the principle of infrared or halogen heating for greater accuracy. These devices determine moisture content by continuously measuring weight loss as the sample is heated. Once weight stabilization is achieved, the moisture percentage is calculated. These machines typically take 5 to 15 minutes to deliver results. However, by the time an out-of-spec moisture level is detected, adjustments may be too late, potentially leading to equipment upsets, production delays, and increased costs. 

In contrast, our approach can deliver accurate moisture readings in seconds. This drastic reduction in processing time would not only enhance the efficiency of the production process but also eliminate critical bottlenecks, leading to smoother operations. As Rahman et al.~\cite{rahman2024comprehensive} demonstrated, even a one-hour delay in MC measurement can result in significant revenue losses. By enabling real-time, rapid MC analysis, our solution can significantly reduce operational downtime, improve resource allocation, and ultimately boost profitability in wood chip-dependent industries.

One notable aspect of the AdaptMoist model architecture is its reliance on straightforward, simple multilayer perceptron (MLP) networks. This choice was primarily driven by the tabular format of the texture features, with a maximum of 63 features in the case of combined features. We found that the MLP network was sufficiently capable of learning from this relatively small feature set, which is why we opted not to use more complex architectures or operations like convolution, max-pooling, or other advanced techniques. A similar approach was applied when selecting the benchmark models. However, a future study could explore how deeper architectures, such as ResNet, DenseNet, and EfficientNet, as well as advanced transformer-based models, might further enhance performance. Additionally, incorporating a broader set of features could also be an interesting direction for such research.

Another potential avenue for future research could involve incorporating raw images directly into the domain adaptation framework. Since Rahman et al.~\cite{rahman2025moistnet} demonstrated promising results using images with their MoistNet architecture, it would be valuable to investigate how raw images could contribute to transferring knowledge and enhancing generalization for wood chip moisture prediction.

%Practical Implications
%Limitations of this study

\section{Conclusions}
%Restating the contribtuions
In this study, we conducted a comprehensive analysis of five texture features for the wood chip MC prediction task. We also relaxed the i.i.d. assumption commonly made in data-driven approaches for wood chip moisture prediction. Specifically, this means that the training and test data can come from different distributions. Under this condition, we found that traditional approaches failed to generalize the performance. Therefore, we proposed AdaptMoist, a domain adaptation method that adversarially learns the shared common features between source and target domains. In this manner the proposed approach can keep the performance consistent on the new unseen target domain (wood chips from new sources). The key takeaways from this study are as follows:
%key takeaways
\begin{itemize}
    \item The integration of all texture features improved the performance of moisture class prediction. For instance, the combined feature set yielded 95\% accuracy compared to the individual best results of 92\%. The discriminative power of each texture feature type contributed collectively to improving the overall accuracy. 
    \item Although the multi-layer perceptron (MLP) achieved the highest accuracy on some individual domains (e.g., S2 and S3) and the second-highest on others (e.g., S1), it struggled to generalize when applied to domain adaptation tasks. In the context of domain adaptation, the source-only model is essentially an MLP network, which faces challenges in transferring knowledge across domains.
    \item Since unsupervised domain adaptation methods lack a direct way to evaluate model performance on the target domain due to the absence of target labels, the AMI callback offers a valuable solution. It enables intermediate evaluation during training with the help of pseudo-labels, allowing for early stopping to prevent overfitting and ensure better generalization.
    \item The proposed AdaptMoist method effectively minimized the domain gap and ensured robust moisture class prediction throughout the domain adaptation tasks. 
\end{itemize}

%Future work
Future work could concentrate on identifying specific Haralick features that enhance transferability and class differentiation for domain adaptation. Further research efforts could explore using deeper architectures, broader feature sets, and the incorporation of raw images to improve domain adaptation and generalization for wood chip moisture prediction. Additionally, with the expanding global wood chip market, deploying the proposed AdaptMoist framework on mobile devices could provide an accessible solution for industries, enabling quick and accurate MC measurement. 

\clearpage
% Start of Appendices
\appendix

\section{Additional Tables}
\setcounter{table}{0}

\begin{table}[h]
\centering
\caption{Performance of machine learning models on S1 images using Haralick features.}
\vspace{0.3cm}
\label{tab:s1_haralick}
\resizebox{\linewidth}{!}{
\begin{tabular}{lcccc}
\hline
Model & Accuracy & Precision & Recall & F1-Score \\
\hline
AdaBoost & 0.6475 (0.0786) & 0.6597 (0.1300) & 0.6475 (0.0786) & 0.6149 (0.1196) \\
Bagging & 0.8700 (0.0242) & 0.8726 (0.0254) & 0.8700 (0.0242) & 0.8699 (0.0242) \\
Decision Tree & 0.8325 (0.0222) & 0.8339 (0.0228) & 0.8325 (0.0222) & 0.8323 (0.0224) \\
Extra Trees & 0.8888 (0.0189) & 0.8893 (0.0194) & 0.8888 (0.0189) & 0.8886 (0.0190) \\
Gradient Boosting & 0.9012 (0.0259) & 0.9018 (0.0259) & 0.9012 (0.0259) & 0.9013 (0.0260) \\
K-Nearest Neighbors & 0.8838 (0.0180) & 0.8852 (0.0167) & 0.8838 (0.0180) & 0.8838 (0.0180) \\
Logistic Regression & 0.9100 (0.0196) & 0.9110 (0.0200) & 0.9100 (0.0196) & 0.9100 (0.0196) \\
Naive Bayes & 0.7475 (0.0104) & 0.7557 (0.0146) & 0.7475 (0.0104) & 0.7366 (0.0115) \\
Neural Network & 0.9175 (0.0155) & 0.9185 (0.0155) & 0.9175 (0.0155) & 0.9174 (0.0153) \\
Random Forest & 0.8850 (0.0147) & 0.8854 (0.0155) & 0.8850 (0.0147) & 0.8848 (0.0146) \\
SVM & 0.9050 (0.0196) & 0.9068 (0.0206) & 0.9050 (0.0196) & 0.9048 (0.0195) \\
Voting Classifier & \textbf{0.9225} (0.0278) & \textbf{0.9233} (0.0281) & \textbf{0.9225} (0.0278) & \textbf{0.9224} (0.0278) \\
\hline
\end{tabular}
}

\end{table}

\begin{table}[h]
\centering
\caption{Performance of machine learning models on S1 images using FOS features.}
\vspace{0.3cm}
\resizebox{\linewidth}{!}{
\begin{tabular}{lcccc}
\hline
Model & Accuracy & Precision & Recall & F1-Score \\
\hline
AdaBoost & 0.6575 (0.0278) & 0.6930 (0.0312) & 0.6575 (0.0278) & 0.6418 (0.0301) \\
Bagging & 0.7063 (0.0317) & 0.7086 (0.0384) & 0.7063 (0.0317) & 0.7052 (0.0343) \\
Decision Tree & 0.6837 (0.0175) & 0.6844 (0.0202) & 0.6837 (0.0175) & 0.6821 (0.0196) \\
Extra Trees & 0.7275 (0.0087) & 0.7268 (0.0129) & 0.7275 (0.0087) & 0.7235 (0.0071) \\
Gradient Boosting & 0.7575 (0.0272) & 0.7572 (0.0302) & 0.7575 (0.0272) & 0.7564 (0.0286) \\
K-Nearest Neighbors & 0.6900 (0.0337) & 0.6864 (0.0355) & 0.6900 (0.0337) & 0.6863 (0.0338) \\
Logistic Regression & 0.7725 (0.0155) & 0.7709 (0.0157) & 0.7725 (0.0155) & 0.7709 (0.0146) \\
Naive Bayes & 0.6813 (0.0206) & 0.6925 (0.0289) & 0.6813 (0.0206) & 0.6667 (0.0198) \\
Neural Network & 0.7738 (0.0278) & 0.7741 (0.0289) & 0.7738 (0.0278) & 0.7724 (0.0273) \\
Random Forest & 0.7625 (0.0260) & 0.7631 (0.0315) & 0.7625 (0.0260) & 0.7596 (0.0271) \\
SVM & 0.7712 (0.0415) & 0.7703 (0.0434) & 0.7712 (0.0415) & 0.7685 (0.0410) \\
Voting Classifier & \textbf{0.7775} (0.0312) & \textbf{0.7768} (0.0347) & \textbf{0.7775} (0.0312) & \textbf{0.7750} (0.0310) \\
\hline
\end{tabular}
}

\end{table}

\begin{table}[h]
\centering
\caption{Performance of machine learning models on S1 images using FPS features.}
\vspace{0.3cm}
\resizebox{\linewidth}{!}{
\begin{tabular}{lcccc}
\hline
Model & Accuracy & Precision & Recall & F1-Score \\
\hline
AdaBoost & 0.5400 (0.0303) & 0.5235 (0.0384) & 0.5400 (0.0303) & 0.5208 (0.0404) \\
Bagging & 0.5212 (0.0357) & 0.5216 (0.0320) & 0.5212 (0.0357) & 0.5199 (0.0347) \\
Decision Tree & 0.5162 (0.0622) & 0.5154 (0.0587) & 0.5162 (0.0622) & 0.5125 (0.0591) \\
Extra Trees & 0.5012 (0.0330) & 0.4945 (0.0285) & 0.5012 (0.0330) & 0.4958 (0.0297) \\
Gradient Boosting & 0.5363 (0.0295) & 0.5236 (0.0256) & 0.5363 (0.0295) & 0.5254 (0.0272) \\
K-Nearest Neighbors & 0.5388 (0.0266) & 0.5373 (0.0243) & 0.5388 (0.0266) & 0.5374 (0.0253) \\
Logistic Regression & 0.5813 (0.0307) & 0.5656 (0.0321) & 0.5813 (0.0307) & 0.5324 (0.0299) \\
Naive Bayes & 0.5837 (0.0266) & 0.5685 (0.0195) & 0.5837 (0.0266) & \textbf{0.5533} (0.0177) \\
Neural Network & 0.5800 (0.0478) & 0.5670 (0.0378) & 0.5800 (0.0478) & 0.5516 (0.0345) \\
Random Forest & 0.5212 (0.0281) & 0.5145 (0.0262) & 0.5212 (0.0281) & 0.5160 (0.0263) \\
SVM & \textbf{0.5925} (0.0413) & \textbf{0.6006} (0.0262) & \textbf{0.5925} (0.0413) & 0.5286 (0.0394) \\
Voting Classifier & 0.5900 (0.0516) & 0.5922 (0.0554) & 0.5900 (0.0516) & 0.5362 (0.0434) \\
\hline
\end{tabular}
}

\end{table}

\begin{table}[h]
\centering
\caption{Performance of machine learning models on S1 images using GLRLM features.}
\vspace{0.3cm}
\resizebox{\linewidth}{!}{
\begin{tabular}{lcccc}
\hline
Model & Accuracy & Precision & Recall & F1-Score \\
\hline
AdaBoost & 0.6775 (0.0185) & 0.6789 (0.0228) & 0.6775 (0.0185) & 0.6608 (0.0194) \\
Bagging & 0.7812 (0.0246) & 0.7823 (0.0244) & 0.7812 (0.0246) & 0.7813 (0.0246) \\
Decision Tree & 0.7488 (0.0309) & 0.7495 (0.0311) & 0.7488 (0.0309) & 0.7487 (0.0310) \\
Extra Trees & 0.7738 (0.0155) & 0.7741 (0.0160) & 0.7738 (0.0155) & 0.7722 (0.0146) \\
Gradient Boosting & 0.8050 (0.0227) & 0.8066 (0.0209) & 0.8050 (0.0227) & 0.8046 (0.0229) \\
K-Nearest Neighbors & 0.7700 (0.0147) & 0.7722 (0.0146) & 0.7700 (0.0147) & 0.7695 (0.0150) \\
Logistic Regression & 0.8037 (0.0347) & 0.8057 (0.0332) & 0.8037 (0.0347) & 0.8039 (0.0348) \\
Naive Bayes & 0.7050 (0.0178) & 0.7113 (0.0216) & 0.7050 (0.0178) & 0.6861 (0.0199) \\
Neural Network & \textbf{0.8087} (0.0239) & \textbf{0.8116} (0.0224) & \textbf{0.8087} (0.0239) & \textbf{0.8086} (0.0245) \\
Random Forest & 0.7825 (0.0275) & 0.7829 (0.0267) & 0.7825 (0.0275) & 0.7821 (0.0274) \\
SVM & 0.8037 (0.0189) & 0.8079 (0.0178) & 0.8037 (0.0189) & 0.8037 (0.0193) \\
Voting Classifier & 0.8063 (0.0232) & 0.8092 (0.0221) & 0.8063 (0.0232) & 0.8063 (0.0236) \\
\hline
\end{tabular}
}

\end{table}

\begin{table}[h]
\centering
\caption{Performance of machine learning models on S1 images using LBP features.}
\vspace{0.3cm}
\resizebox{\linewidth}{!}{
\begin{tabular}{lcccc}
\hline
Model & Accuracy & Precision & Recall & F1-Score \\
\hline
AdaBoost & 0.5938 (0.0607) & 0.5255 (0.1377) & 0.5938 (0.0607) & 0.5302 (0.0816) \\
Bagging & 0.7488 (0.0193) & 0.7513 (0.0187) & 0.7488 (0.0193) & 0.7483 (0.0187) \\
Decision Tree & 0.7275 (0.0206) & 0.7279 (0.0222) & 0.7275 (0.0206) & 0.7260 (0.0200) \\
Extra Trees & 0.7738 (0.0330) & 0.7746 (0.0353) & 0.7738 (0.0330) & 0.7717 (0.0331) \\
Gradient Boosting & 0.7562 (0.0284) & 0.7565 (0.0312) & 0.7562 (0.0284) & 0.7549 (0.0294) \\
K-Nearest Neighbors & 0.7338 (0.0225) & 0.7323 (0.0219) & 0.7338 (0.0225) & 0.7303 (0.0213) \\
Logistic Regression & 0.7688 (0.0450) & 0.7686 (0.0445) & 0.7688 (0.0450) & 0.7670 (0.0446) \\
Naive Bayes & 0.6613 (0.0699) & 0.6549 (0.0790) & 0.6613 (0.0699) & 0.6360 (0.0803) \\
Neural Network & 0.7712 (0.0464) & 0.7717 (0.0458) & 0.7712 (0.0464) & 0.7693 (0.0459) \\
Random Forest & 0.7562 (0.0371) & 0.7553 (0.0397) & 0.7562 (0.0371) & 0.7531 (0.0380) \\
SVM & \textbf{0.7775} (0.0511) & \textbf{0.7768} (0.0518) & \textbf{0.7775} (0.0511) & \textbf{0.7749} (0.0511) \\
Voting Classifier & 0.7712 (0.0464) & 0.7718 (0.0448) & 0.7712 (0.0464) & 0.7699 (0.0458) \\
\hline
\end{tabular}
}

\end{table}

\begin{table}[h]
\centering
\caption{Performance of machine learning models on S1 images using combined features.}
\vspace{0.3cm}
\resizebox{\linewidth}{!}{
\begin{tabular}{lcccc}
\hline
Model & Accuracy & Precision & Recall & F1-Score \\
\hline
AdaBoost & 0.5887 (0.0342) & 0.5701 (0.1074) & 0.5887 (0.0342) & 0.5265 (0.0682) \\
Bagging & 0.9125 (0.0166) & 0.9132 (0.0162) & 0.9125 (0.0166) & 0.9123 (0.0168) \\
Decision Tree & 0.8638 (0.0085) & 0.8637 (0.0099) & 0.8638 (0.0085) & 0.8633 (0.0092) \\
Extra Trees & 0.9187 (0.0165) & 0.9194 (0.0163) & 0.9187 (0.0165) & 0.9187 (0.0162) \\
Gradient Boosting & 0.9250 (0.0191) & 0.9255 (0.0198) & 0.9250 (0.0191) & 0.9250 (0.0193) \\
K-Nearest Neighbors & 0.8975 (0.0194) & 0.8984 (0.0189) & 0.8975 (0.0194) & 0.8974 (0.0192) \\
Logistic Regression & 0.9437 (0.0063) & 0.9444 (0.0067) & 0.9437 (0.0063) & 0.9437 (0.0062) \\
Naive Bayes & 0.7875 (0.0087) & 0.8145 (0.0086) & 0.7875 (0.0087) & 0.7730 (0.0095) \\
Neural Network & 0.9375 (0.0087) & 0.9380 (0.0083) & 0.9375 (0.0087) & 0.9376 (0.0085) \\
Random Forest & 0.9263 (0.0175) & 0.9265 (0.0172) & 0.9263 (0.0175) & 0.9259 (0.0177) \\
SVM & 0.9387 (0.0149) & 0.9394 (0.0141) & 0.9387 (0.0149) & 0.9388 (0.0148) \\
Voting Classifier & \textbf{0.9525} (0.0119) & \textbf{0.9531} (0.0116) & \textbf{0.9525} (0.0119) & \textbf{0.9525} (0.0119) \\
\hline
\end{tabular}
}

\end{table}

\begin{table}[h]
\centering
\caption{Performance of machine learning models on S2 images using Haralick features.}
\vspace{0.3cm}
\resizebox{\linewidth}{!}{
\begin{tabular}{lcccc}
\hline
Model & Accuracy & Precision & Recall & F1-Score \\
\hline
AdaBoost & 0.7975 (0.0371) & 0.8102 (0.0299) & 0.7975 (0.0371) & 0.7994 (0.0366) \\
Bagging & 0.8600 (0.0147) & 0.8630 (0.0140) & 0.8600 (0.0147) & 0.8603 (0.0147) \\
Decision Tree & 0.8237 (0.0246) & 0.8242 (0.0239) & 0.8237 (0.0246) & 0.8224 (0.0244) \\
Extra Trees & 0.8875 (0.0166) & 0.8921 (0.0136) & 0.8875 (0.0166) & 0.8882 (0.0162) \\
Gradient Boosting & 0.9000 (0.0122) & 0.9027 (0.0118) & 0.9000 (0.0122) & 0.9003 (0.0123) \\
K-Nearest Neighbors & 0.8888 (0.0328) & 0.8930 (0.0314) & 0.8888 (0.0328) & 0.8893 (0.0320) \\
Logistic Regression & 0.8975 (0.0065) & 0.9006 (0.0070) & 0.8975 (0.0065) & 0.8979 (0.0068) \\
Naive Bayes & 0.8412 (0.0111) & 0.8455 (0.0114) & 0.8412 (0.0111) & 0.8418 (0.0113) \\
Neural Network & 0.9087 (0.0095) & 0.9102 (0.0092) & 0.9087 (0.0095) & 0.9088 (0.0096) \\
Random Forest & 0.8762 (0.0239) & 0.8820 (0.0193) & 0.8762 (0.0239) & 0.8772 (0.0232) \\
SVM & 0.9050 (0.0158) & 0.9073 (0.0148) & 0.9050 (0.0158) & 0.9054 (0.0156) \\
Voting Classifier & \textbf{0.9150} (0.0173) & \textbf{0.9170} (0.0172) & \textbf{0.9150} (0.0173) & \textbf{0.9151} (0.0173) \\
\hline
\end{tabular}
}

\end{table}

\begin{table}[h]
\centering
\caption{Performance of machine learning models on S2 images using FOS features.}
\vspace{0.3cm}
\resizebox{\linewidth}{!}{
\begin{tabular}{lcccc}
\hline
Model & Accuracy & Precision & Recall & F1-Score \\
\hline
AdaBoost & 0.6238 (0.0214) & 0.6607 (0.0309) & 0.6238 (0.0214) & 0.6025 (0.0278) \\
Bagging & 0.7775 (0.0132) & 0.7821 (0.0160) & 0.7775 (0.0132) & 0.7785 (0.0141) \\
Decision Tree & 0.7050 (0.0147) & 0.7061 (0.0135) & 0.7050 (0.0147) & 0.7052 (0.0143) \\
Extra Trees & 0.7800 (0.0141) & 0.7831 (0.0139) & 0.7800 (0.0141) & 0.7807 (0.0138) \\
Gradient Boosting & 0.7975 (0.0144) & 0.8002 (0.0149) & 0.7975 (0.0144) & 0.7983 (0.0145) \\
K-Nearest Neighbors & 0.7875 (0.0218) & 0.7902 (0.0214) & 0.7875 (0.0218) & 0.7879 (0.0217) \\
Logistic Regression & 0.7888 (0.0210) & 0.7892 (0.0215) & 0.7888 (0.0210) & 0.7884 (0.0208) \\
Naive Bayes & 0.7662 (0.0382) & 0.7655 (0.0384) & 0.7662 (0.0382) & 0.7637 (0.0400) \\
Neural Network & \textbf{0.8025} (0.0119) & \textbf{0.8055} (0.0101) & \textbf{0.8025} (0.0119) & \textbf{0.8033} (0.0114) \\
Random Forest & 0.7925 (0.0185) & 0.7969 (0.0200) & 0.7925 (0.0185) & 0.7933 (0.0184) \\
SVM & 0.7925 (0.0247) & 0.8004 (0.0291) & 0.7925 (0.0247) & 0.7937 (0.0243) \\
Voting Classifier & 0.7988 (0.0125) & 0.8027 (0.0112) & 0.7988 (0.0125) & 0.7995 (0.0117) \\
\hline
\end{tabular}
}

\end{table}

\begin{table}[h]
\centering
\caption{Performance of machine learning models on S2 images using FPS features.}
\vspace{0.3cm}
\resizebox{\linewidth}{!}{
\begin{tabular}{lcccc}
\hline
Model & Accuracy & Precision & Recall & F1-Score \\
\hline
AdaBoost & 0.5988 (0.0735) & 0.6270 (0.0558) & 0.5988 (0.0735) & 0.5834 (0.0883) \\
Bagging & 0.6950 (0.0178) & 0.7020 (0.0140) & 0.6950 (0.0178) & 0.6966 (0.0170) \\
Decision Tree & 0.6512 (0.0405) & 0.6545 (0.0415) & 0.6512 (0.0405) & 0.6516 (0.0408) \\
Extra Trees & 0.6688 (0.0170) & 0.6745 (0.0192) & 0.6688 (0.0170) & 0.6697 (0.0184) \\
Gradient Boosting & 0.6925 (0.0126) & 0.7000 (0.0169) & 0.6925 (0.0126) & 0.6937 (0.0139) \\
K-Nearest Neighbors & 0.7175 (0.0307) & 0.7243 (0.0285) & 0.7175 (0.0307) & 0.7189 (0.0300) \\
Logistic Regression & 0.7212 (0.0275) & 0.7511 (0.0240) & 0.7212 (0.0275) & 0.7230 (0.0276) \\
Naive Bayes & 0.5238 (0.0275) & 0.6071 (0.0131) & 0.5238 (0.0275) & 0.5267 (0.0285) \\
Neural Network & \textbf{0.7325} (0.0301) & \textbf{0.7599} (0.0291) & \textbf{0.7325} (0.0301) & \textbf{0.7337} (0.0294) \\
Random Forest & 0.6925 (0.0096) & 0.6972 (0.0070) & 0.6925 (0.0096) & 0.6934 (0.0089) \\
SVM & 0.6750 (0.0187) & 0.7333 (0.0254) & 0.6750 (0.0187) & 0.6747 (0.0183) \\
Voting Classifier & 0.7087 (0.0155) & 0.7415 (0.0139) & 0.7087 (0.0155) & 0.7097 (0.0153) \\
\hline
\end{tabular}
}

\end{table}

\begin{table}[h]
\centering
\caption{Performance of machine learning models on S2 images using GLRLM features.}
\vspace{0.3cm}
\resizebox{\linewidth}{!}{
\begin{tabular}{lcccc}
\hline
Model & Accuracy & Precision & Recall & F1-Score \\
\hline
AdaBoost & 0.6475 (0.0119) & 0.6780 (0.0111) & 0.6475 (0.0119) & 0.6393 (0.0126) \\
Bagging & 0.8150 (0.0408) & 0.8199 (0.0427) & 0.8150 (0.0408) & 0.8157 (0.0414) \\
Decision Tree & 0.7875 (0.0260) & 0.7908 (0.0284) & 0.7875 (0.0260) & 0.7879 (0.0267) \\
Extra Trees & 0.8150 (0.0196) & 0.8189 (0.0228) & 0.8150 (0.0196) & 0.8154 (0.0204) \\
Gradient Boosting & 0.8200 (0.0283) & 0.8219 (0.0282) & 0.8200 (0.0283) & 0.8201 (0.0286) \\
K-Nearest Neighbors & 0.6987 (0.0371) & 0.7144 (0.0422) & 0.6987 (0.0371) & 0.6998 (0.0370) \\
Logistic Regression & 0.7775 (0.0328) & 0.7980 (0.0317) & 0.7775 (0.0328) & 0.7767 (0.0323) \\
Naive Bayes & 0.4600 (0.0406) & 0.5866 (0.0233) & 0.4600 (0.0406) & 0.4416 (0.0596) \\
Neural Network & 0.7712 (0.0111) & 0.7908 (0.0188) & 0.7712 (0.0111) & 0.7723 (0.0108) \\
Random Forest & \textbf{0.8412} (0.0256) & \textbf{0.8454} (0.0292) & \textbf{0.8412} (0.0256) & \textbf{0.8417} (0.0261) \\
SVM & 0.7438 (0.0155) & 0.8105 (0.0192) & 0.7438 (0.0155) & 0.7427 (0.0162) \\
Voting Classifier & 0.7700 (0.0252) & 0.8034 (0.0225) & 0.7700 (0.0252) & 0.7707 (0.0243) \\
\hline
\end{tabular}
}

\end{table}

\begin{table}[h]
\centering
\caption{Performance of machine learning models on S2 images using LBP features.}
\vspace{0.3cm}
\resizebox{\linewidth}{!}{
\begin{tabular}{lcccc}
\hline
Model & Accuracy & Precision & Recall & F1-Score \\
\hline
AdaBoost & 0.5050 (0.1267) & 0.5185 (0.1184) & 0.5050 (0.1267) & 0.4922 (0.1317) \\
Bagging & 0.7150 (0.0268) & 0.7163 (0.0297) & 0.7150 (0.0268) & 0.7142 (0.0282) \\
Decision Tree & 0.7137 (0.0328) & 0.7147 (0.0328) & 0.7137 (0.0328) & 0.7131 (0.0324) \\
Extra Trees & 0.7338 (0.0229) & 0.7345 (0.0233) & 0.7338 (0.0229) & 0.7329 (0.0237) \\
Gradient Boosting & 0.7162 (0.0287) & 0.7211 (0.0310) & 0.7162 (0.0287) & 0.7155 (0.0317) \\
K-Nearest Neighbors & 0.7375 (0.0377) & 0.7377 (0.0396) & 0.7375 (0.0377) & 0.7357 (0.0394) \\
Logistic Regression & 0.6975 (0.0328) & 0.7020 (0.0319) & 0.6975 (0.0328) & 0.6964 (0.0338) \\
Naive Bayes & 0.6262 (0.0229) & 0.6503 (0.0235) & 0.6262 (0.0229) & 0.6285 (0.0207) \\
Neural Network & 0.7375 (0.0519) & \textbf{0.7440} (0.0500) & 0.7375 (0.0519) & 0.7342 (0.0538) \\
Random Forest & \textbf{0.7400} (0.0294) & 0.7408 (0.0305) & \textbf{0.7400} (0.0294) & \textbf{0.7389} (0.0309) \\
SVM & 0.7163 (0.0357) & 0.7273 (0.0332) & 0.7163 (0.0357) & 0.7146 (0.0361) \\
Voting Classifier & 0.7338 (0.0295) & 0.7402 (0.0257) & 0.7338 (0.0295) & 0.7314 (0.0302) \\
\hline
\end{tabular}
}

\end{table}

\begin{table}[h]
\centering
\caption{Performance of machine learning models on S2 images using combined features.}
\vspace{0.3cm}
\resizebox{\linewidth}{!}{
\begin{tabular}{lcccc}
\hline
Model & Accuracy & Precision & Recall & F1-Score \\
\hline
AdaBoost & 0.7712 (0.0669) & 0.8078 (0.0575) & 0.7712 (0.0669) & 0.7666 (0.0689) \\
Bagging & 0.8838 (0.0202) & 0.8858 (0.0195) & 0.8838 (0.0202) & 0.8840 (0.0199) \\
Decision Tree & 0.8638 (0.0149) & 0.8637 (0.0145) & 0.8638 (0.0149) & 0.8630 (0.0147) \\
Extra Trees & 0.9000 (0.0268) & 0.9033 (0.0240) & 0.9000 (0.0268) & 0.9007 (0.0260) \\
Gradient Boosting & 0.8912 (0.0085) & 0.8935 (0.0063) & 0.8912 (0.0085) & 0.8918 (0.0080) \\
K-Nearest Neighbors & 0.8888 (0.0309) & 0.8907 (0.0299) & 0.8888 (0.0309) & 0.8890 (0.0305) \\
Logistic Regression & 0.9087 (0.0263) & 0.9106 (0.0254) & 0.9087 (0.0263) & 0.9090 (0.0261) \\
Naive Bayes & 0.8362 (0.0149) & 0.8407 (0.0166) & 0.8362 (0.0149) & 0.8364 (0.0155) \\
Neural Network & \textbf{0.9175} (0.0328) & \textbf{0.9185} (0.0324) & \textbf{0.9175} (0.0328) & \textbf{0.9177} (0.0326) \\
Random Forest & 0.8988 (0.0180) & 0.9022 (0.0162) & 0.8988 (0.0180) & 0.8994 (0.0175) \\
SVM & 0.9163 (0.0269) & 0.9183 (0.0258) & 0.9163 (0.0269) & 0.9167 (0.0267) \\
Voting Classifier & 0.9163 (0.0309) & 0.9178 (0.0297) & 0.9163 (0.0309) & 0.9166 (0.0306) \\
\hline
\end{tabular}
}

\end{table}

\begin{table}[h]
\centering
\caption{Performance of machine learning models on S3 images using Haralick features.}
\vspace{0.3cm}
\resizebox{\linewidth}{!}{
\begin{tabular}{lcccc}
\hline
Model & Accuracy & Precision & Recall & F1-Score \\
\hline
AdaBoost & 0.7325 (0.1162) & 0.7271 (0.1644) & 0.7325 (0.1162) & 0.6993 (0.1557) \\
Bagging & 0.8575 (0.0096) & 0.8613 (0.0104) & 0.8575 (0.0096) & 0.8559 (0.0107) \\
Decision Tree & 0.8375 (0.0171) & 0.8388 (0.0188) & 0.8375 (0.0171) & 0.8373 (0.0174) \\
Extra Trees & 0.9000 (0.0316) & 0.9010 (0.0329) & 0.9000 (0.0316) & 0.8993 (0.0322) \\
Gradient Boosting & 0.8825 (0.0263) & 0.8839 (0.0266) & 0.8825 (0.0263) & 0.8825 (0.0266) \\
K-Nearest Neighbors & 0.8775 (0.0250) & 0.8783 (0.0254) & 0.8775 (0.0250) & 0.8761 (0.0262) \\
Logistic Regression & \textbf{0.9500} (0.0141) & \textbf{0.9510} (0.0135) & \textbf{0.9500} (0.0141) & \textbf{0.9498} (0.0143) \\
Naive Bayes & 0.8200 (0.0316) & 0.8372 (0.0411) & 0.8200 (0.0316) & 0.8169 (0.0302) \\
Neural Network & 0.9325 (0.0320) & 0.9330 (0.0318) & 0.9325 (0.0320) & 0.9323 (0.0323) \\
Random Forest & 0.8925 (0.0236) & 0.8927 (0.0232) & 0.8925 (0.0236) & 0.8916 (0.0247) \\
SVM & 0.9175 (0.0222) & 0.9184 (0.0220) & 0.9175 (0.0222) & 0.9170 (0.0227) \\
Voting Classifier & 0.9425 (0.0206) & 0.9442 (0.0197) & 0.9425 (0.0206) & 0.9422 (0.0212) \\
\hline
\end{tabular}
}

\end{table}

\begin{table}[h]
\centering
\caption{Performance of machine learning models on S3 images using FOS features.}
\vspace{0.3cm}
\resizebox{\linewidth}{!}{
\begin{tabular}{lcccc}
\hline
Model & Accuracy & Precision & Recall & F1-Score \\
\hline
AdaBoost & 0.7150 (0.0957) & 0.7176 (0.1492) & 0.7150 (0.0957) & 0.6825 (0.1358) \\
Bagging & 0.8675 (0.0350) & 0.8723 (0.0298) & 0.8675 (0.0350) & 0.8648 (0.0384) \\
Decision Tree & 0.8350 (0.0451) & 0.8376 (0.0476) & 0.8350 (0.0451) & 0.8338 (0.0457) \\
Extra Trees & 0.8600 (0.0216) & 0.8643 (0.0138) & 0.8600 (0.0216) & 0.8571 (0.0260) \\
Gradient Boosting & 0.8700 (0.0469) & 0.8744 (0.0423) & 0.8700 (0.0469) & 0.8671 (0.0504) \\
K-Nearest Neighbors & 0.8325 (0.0250) & 0.8355 (0.0215) & 0.8325 (0.0250) & 0.8295 (0.0285) \\
Logistic Regression & 0.8750 (0.0265) & 0.8785 (0.0223) & 0.8750 (0.0265) & 0.8725 (0.0290) \\
Naive Bayes & 0.7600 (0.0216) & 0.8099 (0.0198) & 0.7600 (0.0216) & 0.7516 (0.0246) \\
Neural Network & 0.8750 (0.0436) & 0.8784 (0.0381) & 0.8750 (0.0436) & 0.8729 (0.0460) \\
Random Forest & 0.8725 (0.0171) & 0.8779 (0.0088) & 0.8725 (0.0171) & 0.8698 (0.0206) \\
SVM & 0.8700 (0.0316) & 0.8766 (0.0216) & 0.8700 (0.0316) & 0.8668 (0.0357) \\
Voting Classifier & \textbf{0.8775} (0.0350) & \textbf{0.8824} (0.0281) & \textbf{0.8775} (0.0350) & \textbf{0.8753} (0.0373) \\
\hline
\end{tabular}
}

\end{table}

\begin{table}[h]
\centering
\caption{Performance of machine learning models on S3 images using FPS features.}
\vspace{0.3cm}
\resizebox{\linewidth}{!}{
\begin{tabular}{lcccc}
\hline
Model & Accuracy & Precision & Recall & F1-Score \\
\hline
AdaBoost & 0.7300 (0.0770) & 0.7691 (0.0641) & 0.7300 (0.0770) & 0.7203 (0.0820) \\
Bagging & 0.7700 (0.0469) & 0.7738 (0.0462) & 0.7700 (0.0469) & 0.7674 (0.0465) \\
Decision Tree & 0.7650 (0.0265) & 0.7663 (0.0264) & 0.7650 (0.0265) & 0.7635 (0.0263) \\
Extra Trees & 0.7750 (0.0265) & 0.7751 (0.0251) & 0.7750 (0.0265) & 0.7727 (0.0257) \\
Gradient Boosting & 0.7900 (0.0316) & 0.7923 (0.0277) & 0.7900 (0.0316) & 0.7871 (0.0322) \\
K-Nearest Neighbors & 0.7800 (0.0365) & 0.7817 (0.0397) & 0.7800 (0.0365) & 0.7759 (0.0395) \\
Logistic Regression & 0.7875 (0.0171) & 0.7897 (0.0195) & 0.7875 (0.0171) & 0.7837 (0.0194) \\
Naive Bayes & 0.8000 (0.0432) & 0.8135 (0.0399) & 0.8000 (0.0432) & 0.7949 (0.0473) \\
Neural Network & 0.8000 (0.0337) & 0.8082 (0.0341) & 0.8000 (0.0337) & 0.7960 (0.0355) \\
Random Forest & 0.7825 (0.0330) & 0.7845 (0.0276) & 0.7825 (0.0330) & 0.7803 (0.0322) \\
SVM & \textbf{0.8100} (0.0346) & \textbf{0.8271} (0.0244) & \textbf{0.8100} (0.0346) & \textbf{0.8034} (0.0391) \\
Voting Classifier & 0.7975 (0.0386) & 0.8036 (0.0361) & 0.7975 (0.0386) & 0.7934 (0.0415) \\
\hline
\end{tabular}
}

\end{table}

\begin{table}[h]
\centering
\caption{Performance of machine learning models on S3 images using GLRLM features.}
\vspace{0.3cm}
\resizebox{\linewidth}{!}{
\begin{tabular}{lcccc}
\hline
Model & Accuracy & Precision & Recall & F1-Score \\
\hline
AdaBoost & 0.7200 (0.1023) & 0.7891 (0.0908) & 0.7200 (0.1023) & 0.7061 (0.1122) \\
Bagging & 0.8175 (0.0275) & 0.8211 (0.0321) & 0.8175 (0.0275) & 0.8135 (0.0268) \\
Decision Tree & 0.8000 (0.0462) & 0.8067 (0.0459) & 0.8000 (0.0462) & 0.7986 (0.0478) \\
Extra Trees & 0.8125 (0.0386) & 0.8134 (0.0397) & 0.8125 (0.0386) & 0.8076 (0.0404) \\
Gradient Boosting & 0.8150 (0.0129) & 0.8178 (0.0159) & 0.8150 (0.0129) & 0.8125 (0.0135) \\
K-Nearest Neighbors & 0.8125 (0.0150) & 0.8174 (0.0207) & 0.8125 (0.0150) & 0.8062 (0.0150) \\
Logistic Regression & 0.8300 (0.0216) & 0.8328 (0.0214) & 0.8300 (0.0216) & 0.8265 (0.0242) \\
Naive Bayes & \textbf{0.8400} (0.0216) & 0.8451 (0.0186) & \textbf{0.8400} (0.0216) & \textbf{0.8352} (0.0252) \\
Neural Network & 0.8225 (0.0419) & 0.8235 (0.0415) & 0.8225 (0.0419) & 0.8188 (0.0433) \\
Random Forest & 0.8275 (0.0171) & 0.8321 (0.0173) & 0.8275 (0.0171) & 0.8233 (0.0178) \\
SVM & \textbf{0.8400} (0.0294) & \textbf{0.8545} (0.0215) & \textbf{0.8400} (0.0294) & 0.8317 (0.0341) \\
Voting Classifier & 0.8175 (0.0171) & 0.8194 (0.0198) & 0.8175 (0.0171) & 0.8128 (0.0182) \\
\hline
\end{tabular}
}

\end{table}

\begin{table}[h]
\centering
\caption{Performance of machine learning models on S3 images using LBP features.}
\vspace{0.3cm}
\resizebox{\linewidth}{!}{
\begin{tabular}{lcccc}
\hline
Model & Accuracy & Precision & Recall & F1-Score \\
\hline
AdaBoost & 0.6650 (0.0480) & 0.6856 (0.0422) & 0.6650 (0.0480) & 0.6563 (0.0502) \\
Bagging & 0.7750 (0.0465) & 0.7757 (0.0472) & 0.7750 (0.0465) & 0.7743 (0.0474) \\
Decision Tree & 0.7200 (0.0469) & 0.7241 (0.0462) & 0.7200 (0.0469) & 0.7182 (0.0468) \\
Extra Trees & 0.7775 (0.0299) & 0.7801 (0.0337) & 0.7775 (0.0299) & 0.7774 (0.0319) \\
Gradient Boosting & 0.7650 (0.0480) & 0.7661 (0.0478) & 0.7650 (0.0480) & 0.7649 (0.0481) \\
K-Nearest Neighbors & 0.7750 (0.0614) & 0.7807 (0.0625) & 0.7750 (0.0614) & 0.7747 (0.0621) \\
Logistic Regression & 0.8050 (0.0173) & 0.8082 (0.0178) & 0.8050 (0.0173) & 0.8029 (0.0172) \\
Naive Bayes & 0.4550 (0.0420) & 0.4697 (0.0388) & 0.4550 (0.0420) & 0.4302 (0.0360) \\
Neural Network & \textbf{0.8150} (0.0289) & \textbf{0.8195} (0.0317) & \textbf{0.8150} (0.0289) & \textbf{0.8145} (0.0300) \\
Random Forest & 0.7850 (0.0436) & 0.7885 (0.0446) & 0.7850 (0.0436) & 0.7845 (0.0446) \\
SVM & 0.7675 (0.0250) & 0.7717 (0.0265) & 0.7675 (0.0250) & 0.7668 (0.0259) \\
Voting Classifier & 0.8050 (0.0289) & 0.8110 (0.0306) & 0.8050 (0.0289) & 0.8045 (0.0294) \\
\hline
\end{tabular}
}

\end{table}

\begin{table}[h]
\centering
\caption{Performance of machine learning models on S3 images using combined features.}
\vspace{0.3cm}
\resizebox{\linewidth}{!}{
\begin{tabular}{lcccc}
\hline
Model & Accuracy & Precision & Recall & F1-Score \\
\hline
AdaBoost & 0.7575 (0.1276) & 0.7474 (0.1745) & 0.7575 (0.1276) & 0.7345 (0.1707) \\
Bagging & 0.8750 (0.0311) & 0.8805 (0.0270) & 0.8750 (0.0311) & 0.8727 (0.0337) \\
Decision Tree & 0.8400 (0.0082) & 0.8449 (0.0103) & 0.8400 (0.0082) & 0.8396 (0.0069) \\
Extra Trees & 0.8900 (0.0374) & 0.8941 (0.0319) & 0.8900 (0.0374) & 0.8876 (0.0409) \\
Gradient Boosting & 0.8850 (0.0300) & 0.8866 (0.0287) & 0.8850 (0.0300) & 0.8841 (0.0312) \\
K-Nearest Neighbors & 0.8750 (0.0129) & 0.8767 (0.0138) & 0.8750 (0.0129) & 0.8744 (0.0138) \\
Logistic Regression & 0.9475 (0.0126) & 0.9480 (0.0120) & 0.9475 (0.0126) & 0.9473 (0.0128) \\
Naive Bayes & 0.8250 (0.0100) & 0.8522 (0.0121) & 0.8250 (0.0100) & 0.8206 (0.0117) \\
Neural Network & \textbf{0.9500} (0.0183) & \textbf{0.9509} (0.0174) & \textbf{0.9500} (0.0183) & \textbf{0.9497} (0.0186) \\
Random Forest & 0.8675 (0.0171) & 0.8699 (0.0139) & 0.8675 (0.0171) & 0.8653 (0.0192) \\
SVM & 0.9150 (0.0387) & 0.9189 (0.0327) & 0.9150 (0.0387) & 0.9131 (0.0415) \\
Voting Classifier & 0.9400 (0.0216) & 0.9409 (0.0208) & 0.9400 (0.0216) & 0.9396 (0.0219) \\
\hline
\end{tabular}
}

\end{table}

\begin{table}[h]
\centering
\caption{Performance of machine learning models on S2-B1 images using Haralick features.}
\vspace{0.3cm}
\resizebox{\linewidth}{!}{
\begin{tabular}{lcccc}
\hline
Model & Accuracy & Precision & Recall & F1-Score \\
\hline
AdaBoost & 0.7475 (0.0737) & 0.7689 (0.0601) & 0.7475 (0.0737) & 0.7400 (0.0821) \\
Bagging & 0.8825 (0.0250) & 0.8872 (0.0238) & 0.8825 (0.0250) & 0.8824 (0.0253) \\
Decision Tree & 0.8425 (0.0275) & 0.8437 (0.0269) & 0.8425 (0.0275) & 0.8427 (0.0271) \\
Extra Trees & 0.9000 (0.0163) & 0.9047 (0.0159) & 0.9000 (0.0163) & 0.9001 (0.0161) \\
Gradient Boosting & 0.9050 (0.0311) & 0.9110 (0.0246) & 0.9050 (0.0311) & 0.9050 (0.0308) \\
K-Nearest Neighbors & 0.9025 (0.0250) & 0.9053 (0.0277) & 0.9025 (0.0250) & 0.9026 (0.0251) \\
Logistic Regression & 0.8875 (0.0369) & 0.8896 (0.0362) & 0.8875 (0.0369) & 0.8876 (0.0372) \\
Naive Bayes & 0.8375 (0.0275) & 0.8423 (0.0292) & 0.8375 (0.0275) & 0.8379 (0.0279) \\
Neural Network & \textbf{0.9200} (0.0258) & \textbf{0.9217} (0.0257) & \textbf{0.9200} (0.0258) & \textbf{0.9201} (0.0260) \\
Random Forest & 0.8900 (0.0082) & 0.8953 (0.0102) & 0.8900 (0.0082) & 0.8902 (0.0086) \\
SVM & 0.9050 (0.0289) & 0.9082 (0.0279) & 0.9050 (0.0289) & 0.9053 (0.0286) \\
Voting Classifier & 0.9150 (0.0173) & 0.9180 (0.0158) & 0.9150 (0.0173) & 0.9152 (0.0174) \\
\hline
\end{tabular}
}

\end{table}

\begin{table}[h]
\centering
\caption{Performance of machine learning models on S2-B1 images using FOS features.}
\vspace{0.3cm}
\resizebox{\linewidth}{!}{
\begin{tabular}{lcccc}
\hline
Model & Accuracy & Precision & Recall & F1-Score \\
\hline
AdaBoost & 0.6200 (0.1068) & 0.6417 (0.0937) & 0.6200 (0.1068) & 0.5986 (0.1290) \\
Bagging & 0.7850 (0.0173) & 0.7902 (0.0220) & 0.7850 (0.0173) & 0.7845 (0.0189) \\
Decision Tree & 0.7250 (0.0480) & 0.7274 (0.0500) & 0.7250 (0.0480) & 0.7232 (0.0465) \\
Extra Trees & 0.7675 (0.0150) & 0.7744 (0.0179) & 0.7675 (0.0150) & 0.7667 (0.0168) \\
Gradient Boosting & 0.7675 (0.0222) & 0.7775 (0.0281) & 0.7675 (0.0222) & 0.7681 (0.0242) \\
K-Nearest Neighbors & 0.8050 (0.0265) & 0.8076 (0.0257) & 0.8050 (0.0265) & 0.8051 (0.0265) \\
Logistic Regression & 0.8050 (0.0265) & 0.8084 (0.0275) & 0.8050 (0.0265) & 0.8033 (0.0287) \\
Naive Bayes & 0.7600 (0.0141) & 0.7584 (0.0140) & 0.7600 (0.0141) & 0.7573 (0.0137) \\
Neural Network & \textbf{0.8150} (0.0252) & \textbf{0.8187} (0.0249) & \textbf{0.8150} (0.0252) & \textbf{0.8141} (0.0246) \\
Random Forest & 0.7800 (0.0200) & 0.7842 (0.0244) & 0.7800 (0.0200) & 0.7785 (0.0212) \\
SVM & 0.7950 (0.0238) & 0.8034 (0.0281) & 0.7950 (0.0238) & 0.7953 (0.0231) \\
Voting Classifier & 0.8025 (0.0330) & 0.8073 (0.0323) & 0.8025 (0.0330) & 0.8015 (0.0325) \\
\hline
\end{tabular}
}

\end{table}

\begin{table}[h]
\centering
\caption{Performance of machine learning models on S2-B1 images using FPS features.}
\vspace{0.3cm}
\resizebox{\linewidth}{!}{
\begin{tabular}{lcccc}
\hline
Model & Accuracy & Precision & Recall & F1-Score \\
\hline
AdaBoost & 0.5775 (0.0512) & 0.6249 (0.0497) & 0.5775 (0.0512) & 0.5562 (0.0619) \\
Bagging & 0.6400 (0.0535) & 0.6446 (0.0492) & 0.6400 (0.0535) & 0.6387 (0.0530) \\
Decision Tree & 0.6225 (0.0556) & 0.6279 (0.0557) & 0.6225 (0.0556) & 0.6225 (0.0549) \\
Extra Trees & 0.6350 (0.0370) & 0.6447 (0.0296) & 0.6350 (0.0370) & 0.6343 (0.0342) \\
Gradient Boosting & 0.6900 (0.0365) & 0.6931 (0.0357) & 0.6900 (0.0365) & 0.6900 (0.0368) \\
K-Nearest Neighbors & \textbf{0.7275} (0.0222) & 0.7365 (0.0156) & \textbf{0.7275} (0.0222) & \textbf{0.7280} (0.0211) \\
Logistic Regression & 0.6950 (0.0420) & 0.7219 (0.0301) & 0.6950 (0.0420) & 0.6965 (0.0415) \\
Naive Bayes & 0.5050 (0.0823) & 0.6073 (0.0212) & 0.5050 (0.0823) & 0.5000 (0.0883) \\
Neural Network & 0.7150 (0.0238) & \textbf{0.7515} (0.0137) & 0.7150 (0.0238) & 0.7156 (0.0240) \\
Random Forest & 0.6875 (0.0275) & 0.6944 (0.0283) & 0.6875 (0.0275) & 0.6869 (0.0272) \\
SVM & 0.6675 (0.0320) & 0.7125 (0.0204) & 0.6675 (0.0320) & 0.6673 (0.0329) \\
Voting Classifier & 0.6725 (0.0222) & 0.7062 (0.0097) & 0.6725 (0.0222) & 0.6729 (0.0218) \\
\hline
\end{tabular}
}

\end{table}

\begin{table}[h]
\centering
\caption{Performance of machine learning models on S2-B1 images using GLRLM features.}
\vspace{0.3cm}
\resizebox{\linewidth}{!}{
\begin{tabular}{lcccc}
\hline
Model & Accuracy & Precision & Recall & F1-Score \\
\hline
AdaBoost & 0.6100 (0.0535) & 0.6583 (0.0529) & 0.6100 (0.0535) & 0.5823 (0.0557) \\
Bagging & 0.7775 (0.0550) & 0.7856 (0.0565) & 0.7775 (0.0550) & 0.7769 (0.0559) \\
Decision Tree & 0.7375 (0.0222) & 0.7410 (0.0204) & 0.7375 (0.0222) & 0.7363 (0.0242) \\
Extra Trees & \textbf{0.7875} (0.0340) & 0.7988 (0.0374) & \textbf{0.7875} (0.0340) & \textbf{0.7883} (0.0337) \\
Gradient Boosting & 0.7425 (0.0150) & 0.7511 (0.0141) & 0.7425 (0.0150) & 0.7422 (0.0149) \\
K-Nearest Neighbors & 0.6725 (0.0171) & 0.6939 (0.0192) & 0.6725 (0.0171) & 0.6742 (0.0169) \\
Logistic Regression & 0.7625 (0.0350) & 0.7895 (0.0356) & 0.7625 (0.0350) & 0.7620 (0.0355) \\
Naive Bayes & 0.4750 (0.0436) & 0.5844 (0.0572) & 0.4750 (0.0436) & 0.4673 (0.0560) \\
Neural Network & 0.7550 (0.0238) & 0.7909 (0.0083) & 0.7550 (0.0238) & 0.7548 (0.0248) \\
Random Forest & 0.7850 (0.0265) & 0.7976 (0.0171) & 0.7850 (0.0265) & 0.7858 (0.0253) \\
SVM & 0.7225 (0.0457) & \textbf{0.8026} (0.0125) & 0.7225 (0.0457) & 0.7167 (0.0522) \\
Voting Classifier & 0.7600 (0.0294) & 0.7962 (0.0205) & 0.7600 (0.0294) & 0.7600 (0.0290) \\
\hline
\end{tabular}
}

\end{table}

\begin{table}[h]
\centering
\caption{Performance of machine learning models on S2-B1 images using LBP features.}
\vspace{0.3cm}
\resizebox{\linewidth}{!}{
\begin{tabular}{lcccc}
\hline
Model & Accuracy & Precision & Recall & F1-Score \\
\hline
AdaBoost & 0.6100 (0.0294) & 0.6120 (0.0330) & 0.6100 (0.0294) & 0.6024 (0.0381) \\
Bagging & 0.7525 (0.0236) & 0.7539 (0.0230) & 0.7525 (0.0236) & 0.7481 (0.0221) \\
Decision Tree & 0.7125 (0.0330) & 0.7133 (0.0355) & 0.7125 (0.0330) & 0.7102 (0.0328) \\
Extra Trees & 0.7700 (0.0216) & 0.7734 (0.0217) & 0.7700 (0.0216) & 0.7705 (0.0212) \\
Gradient Boosting & 0.7425 (0.0299) & 0.7456 (0.0242) & 0.7425 (0.0299) & 0.7422 (0.0274) \\
K-Nearest Neighbors & \textbf{0.7825} (0.0492) & \textbf{0.7873} (0.0483) & \textbf{0.7825} (0.0492) & \textbf{0.7832} (0.0508) \\
Logistic Regression & 0.7125 (0.0171) & 0.7180 (0.0139) & 0.7125 (0.0171) & 0.7054 (0.0163) \\
Naive Bayes & 0.6500 (0.0416) & 0.6745 (0.0361) & 0.6500 (0.0416) & 0.6504 (0.0425) \\
Neural Network & 0.7325 (0.0330) & 0.7358 (0.0301) & 0.7325 (0.0330) & 0.7271 (0.0302) \\
Random Forest & 0.7500 (0.0455) & 0.7546 (0.0429) & 0.7500 (0.0455) & 0.7491 (0.0425) \\
SVM & 0.7100 (0.0294) & 0.7218 (0.0374) & 0.7100 (0.0294) & 0.7071 (0.0299) \\
Voting Classifier & 0.7150 (0.0129) & 0.7188 (0.0098) & 0.7150 (0.0129) & 0.7078 (0.0121) \\
\hline
\end{tabular}
}

\end{table}

\begin{table}[h]
\centering
\caption{Performance of machine learning models on S2-B1 images using combined features.}
\vspace{0.3cm}
\resizebox{\linewidth}{!}{
\begin{tabular}{lcccc}
\hline
Model & Accuracy & Precision & Recall & F1-Score \\
\hline
AdaBoost & 0.7750 (0.1204) & 0.8014 (0.1013) & 0.7750 (0.1204) & 0.7716 (0.1237) \\
Bagging & 0.8950 (0.0379) & 0.9000 (0.0385) & 0.8950 (0.0379) & 0.8948 (0.0384) \\
Decision Tree & 0.8300 (0.0497) & 0.8365 (0.0470) & 0.8300 (0.0497) & 0.8308 (0.0494) \\
Extra Trees & 0.8925 (0.0171) & 0.8979 (0.0150) & 0.8925 (0.0171) & 0.8925 (0.0165) \\
Gradient Boosting & 0.9025 (0.0126) & 0.9052 (0.0105) & 0.9025 (0.0126) & 0.9024 (0.0122) \\
K-Nearest Neighbors & 0.9050 (0.0208) & 0.9065 (0.0222) & 0.9050 (0.0208) & 0.9049 (0.0205) \\
Logistic Regression & \textbf{0.9300} (0.0294) & \textbf{0.9313} (0.0282) & \textbf{0.9300} (0.0294) & \textbf{0.9301} (0.0291) \\
Naive Bayes & 0.8325 (0.0050) & 0.8348 (0.0048) & 0.8325 (0.0050) & 0.8324 (0.0046) \\
Neural Network & 0.9250 (0.0173) & 0.9263 (0.0166) & 0.9250 (0.0173) & 0.9250 (0.0174) \\
Random Forest & 0.8875 (0.0263) & 0.8937 (0.0271) & 0.8875 (0.0263) & 0.8876 (0.0264) \\
SVM & 0.9050 (0.0129) & 0.9084 (0.0126) & 0.9050 (0.0129) & 0.9054 (0.0130) \\
Voting Classifier & 0.9225 (0.0096) & 0.9238 (0.0083) & 0.9225 (0.0096) & 0.9225 (0.0093) \\
\hline
\end{tabular}
}

\end{table}

\begin{table}[h]
\centering
\caption{Performance of machine learning models on S2-B2 images using Haralick features.}
\vspace{0.3cm}
\resizebox{\linewidth}{!}{
\begin{tabular}{lcccc}
\hline
Model & Accuracy & Precision & Recall & F1-Score \\
\hline
AdaBoost & 0.7650 (0.0933) & 0.8189 (0.0475) & 0.7650 (0.0933) & 0.7576 (0.1002) \\
Bagging & 0.8550 (0.0332) & 0.8582 (0.0339) & 0.8550 (0.0332) & 0.8553 (0.0340) \\
Decision Tree & 0.8225 (0.0435) & 0.8241 (0.0448) & 0.8225 (0.0435) & 0.8222 (0.0436) \\
Extra Trees & 0.8775 (0.0263) & 0.8853 (0.0264) & 0.8775 (0.0263) & 0.8777 (0.0251) \\
Gradient Boosting & 0.8750 (0.0191) & 0.8771 (0.0207) & 0.8750 (0.0191) & 0.8747 (0.0187) \\
K-Nearest Neighbors & 0.9000 (0.0365) & 0.9063 (0.0342) & 0.9000 (0.0365) & 0.9003 (0.0360) \\
Logistic Regression & 0.9025 (0.0206) & 0.9044 (0.0211) & 0.9025 (0.0206) & 0.9028 (0.0208) \\
Naive Bayes & 0.8500 (0.0163) & 0.8564 (0.0149) & 0.8500 (0.0163) & 0.8492 (0.0175) \\
Neural Network & 0.9200 (0.0183) & 0.9219 (0.0172) & 0.9200 (0.0183) & 0.9202 (0.0182) \\
Random Forest & 0.8675 (0.0171) & 0.8723 (0.0158) & 0.8675 (0.0171) & 0.8671 (0.0164) \\
SVM & 0.9100 (0.0141) & 0.9162 (0.0092) & 0.9100 (0.0141) & 0.9097 (0.0144) \\
Voting Classifier & \textbf{0.9275} (0.0096) & \textbf{0.9312} (0.0076) & \textbf{0.9275} (0.0096) & \textbf{0.9275} (0.0097) \\
\hline
\end{tabular}
}

\end{table}

\begin{table}[h]
\centering
\caption{Performance of machine learning models on S2-B2 images using FOS features.}
\vspace{0.3cm}
\resizebox{\linewidth}{!}{
\begin{tabular}{lcccc}
\hline
Model & Accuracy & Precision & Recall & F1-Score \\
\hline
AdaBoost & 0.6100 (0.0909) & 0.7134 (0.0383) & 0.6100 (0.0909) & 0.5851 (0.1038) \\
Bagging & 0.7475 (0.0287) & 0.7539 (0.0306) & 0.7475 (0.0287) & 0.7480 (0.0275) \\
Decision Tree & 0.7250 (0.0420) & 0.7279 (0.0440) & 0.7250 (0.0420) & 0.7205 (0.0451) \\
Extra Trees & \textbf{0.8050} (0.0173) & 0.8131 (0.0138) & \textbf{0.8050} (0.0173) & \textbf{0.8037} (0.0146) \\
Gradient Boosting & 0.7900 (0.0337) & 0.7963 (0.0260) & 0.7900 (0.0337) & 0.7897 (0.0303) \\
K-Nearest Neighbors & 0.7675 (0.0299) & 0.7695 (0.0257) & 0.7675 (0.0299) & 0.7658 (0.0268) \\
Logistic Regression & 0.7850 (0.0129) & 0.7909 (0.0129) & 0.7850 (0.0129) & 0.7853 (0.0116) \\
Naive Bayes & 0.7575 (0.0377) & 0.7624 (0.0376) & 0.7575 (0.0377) & 0.7562 (0.0375) \\
Neural Network & 0.7850 (0.0342) & 0.7885 (0.0296) & 0.7850 (0.0342) & 0.7850 (0.0322) \\
Random Forest & 0.7775 (0.0222) & 0.7830 (0.0184) & 0.7775 (0.0222) & 0.7764 (0.0208) \\
SVM & 0.7925 (0.0512) & \textbf{0.8134} (0.0484) & 0.7925 (0.0512) & 0.7901 (0.0483) \\
Voting Classifier & 0.7975 (0.0465) & 0.8024 (0.0420) & 0.7975 (0.0465) & 0.7971 (0.0444) \\
\hline
\end{tabular}
}

\end{table}

\begin{table}[h]
\centering
\caption{Performance of machine learning models on S2-B2 images using FPS features.}
\vspace{0.3cm}
\resizebox{\linewidth}{!}{
\begin{tabular}{lcccc}
\hline
Model & Accuracy & Precision & Recall & F1-Score \\
\hline
AdaBoost & 0.5550 (0.1282) & 0.6151 (0.1125) & 0.5550 (0.1282) & 0.5458 (0.1402) \\
Bagging & 0.6700 (0.0216) & 0.6721 (0.0210) & 0.6700 (0.0216) & 0.6687 (0.0221) \\
Decision Tree & 0.6475 (0.0465) & 0.6460 (0.0463) & 0.6475 (0.0465) & 0.6450 (0.0468) \\
Extra Trees & 0.6725 (0.0287) & 0.6742 (0.0295) & 0.6725 (0.0287) & 0.6713 (0.0300) \\
Gradient Boosting & 0.6875 (0.0386) & 0.6963 (0.0505) & 0.6875 (0.0386) & 0.6881 (0.0396) \\
K-Nearest Neighbors & 0.6400 (0.0600) & 0.6412 (0.0606) & 0.6400 (0.0600) & 0.6395 (0.0600) \\
Logistic Regression & 0.7025 (0.0275) & 0.7434 (0.0533) & 0.7025 (0.0275) & 0.7025 (0.0271) \\
Naive Bayes & 0.5250 (0.0772) & 0.6016 (0.0267) & 0.5250 (0.0772) & 0.5264 (0.0792) \\
Neural Network & \textbf{0.7125} (0.0171) & \textbf{0.7560} (0.0361) & \textbf{0.7125} (0.0171) & \textbf{0.7123} (0.0182) \\
Random Forest & 0.6700 (0.0183) & 0.6697 (0.0204) & 0.6700 (0.0183) & 0.6681 (0.0191) \\
SVM & 0.6750 (0.0129) & 0.7389 (0.0380) & 0.6750 (0.0129) & 0.6719 (0.0153) \\
Voting Classifier & 0.6975 (0.0206) & 0.7390 (0.0357) & 0.6975 (0.0206) & 0.6975 (0.0207) \\
\hline
\end{tabular}
}

\end{table}

\begin{table}[h]
\centering
\caption{Performance of machine learning models on S2-B2 images using GLRLM features.}
\vspace{0.3cm}
\resizebox{\linewidth}{!}{
\begin{tabular}{lcccc}
\hline
Model & Accuracy & Precision & Recall & F1-Score \\
\hline
AdaBoost & 0.6825 (0.0171) & 0.7199 (0.0352) & 0.6825 (0.0171) & 0.6775 (0.0139) \\
Bagging & \textbf{0.8275} (0.0602) & \textbf{0.8297} (0.0586) & \textbf{0.8275} (0.0602) & \textbf{0.8279} (0.0593) \\
Decision Tree & 0.7975 (0.0222) & 0.8002 (0.0251) & 0.7975 (0.0222) & 0.7976 (0.0234) \\
Extra Trees & 0.8025 (0.0759) & 0.8042 (0.0777) & 0.8025 (0.0759) & 0.8023 (0.0763) \\
Gradient Boosting & 0.8150 (0.0420) & 0.8193 (0.0429) & 0.8150 (0.0420) & 0.8154 (0.0410) \\
K-Nearest Neighbors & 0.6775 (0.0465) & 0.6940 (0.0453) & 0.6775 (0.0465) & 0.6797 (0.0446) \\
Logistic Regression & 0.7625 (0.0359) & 0.7791 (0.0398) & 0.7625 (0.0359) & 0.7627 (0.0376) \\
Naive Bayes & 0.4425 (0.0411) & 0.5104 (0.1407) & 0.4425 (0.0411) & 0.4150 (0.0652) \\
Neural Network & 0.7600 (0.0383) & 0.7896 (0.0316) & 0.7600 (0.0383) & 0.7603 (0.0384) \\
Random Forest & 0.8225 (0.0492) & 0.8256 (0.0508) & 0.8225 (0.0492) & 0.8212 (0.0489) \\
SVM & 0.7550 (0.0289) & 0.8196 (0.0321) & 0.7550 (0.0289) & 0.7515 (0.0301) \\
Voting Classifier & 0.7800 (0.0408) & 0.8229 (0.0329) & 0.7800 (0.0408) & 0.7796 (0.0414) \\
\hline
\end{tabular}
}

\end{table}

\begin{table}[h]
\centering
\caption{Performance of machine learning models on S2-B2 images using LBP features.}
\vspace{0.3cm}
\resizebox{\linewidth}{!}{
\begin{tabular}{lcccc}
\hline
Model & Accuracy & Precision & Recall & F1-Score \\
\hline
AdaBoost & 0.5175 (0.0991) & 0.5148 (0.0970) & 0.5175 (0.0991) & 0.4981 (0.1069) \\
Bagging & 0.7400 (0.0748) & 0.7424 (0.0767) & 0.7400 (0.0748) & 0.7392 (0.0756) \\
Decision Tree & 0.7025 (0.0562) & 0.7013 (0.0614) & 0.7025 (0.0562) & 0.7005 (0.0595) \\
Extra Trees & \textbf{0.7650} (0.0342) & 0.7676 (0.0361) & \textbf{0.7650} (0.0342) & \textbf{0.7645} (0.0349) \\
Gradient Boosting & 0.7225 (0.0499) & 0.7324 (0.0537) & 0.7225 (0.0499) & 0.7235 (0.0503) \\
K-Nearest Neighbors & 0.7575 (0.0411) & 0.7561 (0.0424) & 0.7575 (0.0411) & 0.7564 (0.0421) \\
Logistic Regression & 0.7075 (0.0685) & 0.7212 (0.0701) & 0.7075 (0.0685) & 0.7095 (0.0664) \\
Naive Bayes & 0.6350 (0.0332) & 0.6566 (0.0230) & 0.6350 (0.0332) & 0.6372 (0.0337) \\
Neural Network & 0.7350 (0.0624) & 0.7559 (0.0604) & 0.7350 (0.0624) & 0.7359 (0.0622) \\
Random Forest & 0.7450 (0.0420) & 0.7470 (0.0418) & 0.7450 (0.0420) & 0.7432 (0.0421) \\
SVM & 0.7350 (0.0465) & \textbf{0.7929} (0.0445) & 0.7350 (0.0465) & 0.7356 (0.0454) \\
Voting Classifier & 0.7350 (0.0597) & 0.7644 (0.0575) & 0.7350 (0.0597) & 0.7366 (0.0587) \\
\hline
\end{tabular}
}

\end{table}

\begin{table}[h]
\centering
\caption{Performance of machine learning models on S2-B2 images using combined features.}
\label{tab:s2b2_combined}
\vspace{0.3cm}
\resizebox{\linewidth}{!}{
\begin{tabular}{lcccc}
\hline
Model & Accuracy & Precision & Recall & F1-Score \\
\hline
AdaBoost & 0.7900 (0.1120) & 0.8394 (0.0472) & 0.7900 (0.1120) & 0.7835 (0.1201) \\
Bagging & 0.8700 (0.0216) & 0.8724 (0.0213) & 0.8700 (0.0216) & 0.8699 (0.0215) \\
Decision Tree & 0.8800 (0.0356) & 0.8804 (0.0356) & 0.8800 (0.0356) & 0.8800 (0.0355) \\
Extra Trees & 0.8950 (0.0252) & 0.8986 (0.0232) & 0.8950 (0.0252) & 0.8951 (0.0246) \\
Gradient Boosting & 0.9000 (0.0271) & 0.9022 (0.0283) & 0.9000 (0.0271) & 0.8997 (0.0273) \\
K-Nearest Neighbors & 0.8950 (0.0129) & 0.9000 (0.0124) & 0.8950 (0.0129) & 0.8951 (0.0128) \\
Logistic Regression & 0.9225 (0.0150) & 0.9255 (0.0143) & 0.9225 (0.0150) & 0.9226 (0.0146) \\
Naive Bayes & 0.8525 (0.0340) & 0.8621 (0.0376) & 0.8525 (0.0340) & 0.8509 (0.0331) \\
Neural Network & 0.9225 (0.0206) & 0.9240 (0.0212) & 0.9225 (0.0206) & 0.9226 (0.0206) \\
Random Forest & 0.8775 (0.0189) & 0.8837 (0.0204) & 0.8775 (0.0189) & 0.8771 (0.0178) \\
SVM & 0.9125 (0.0096) & 0.9179 (0.0068) & 0.9125 (0.0096) & 0.9121 (0.0093) \\
Voting Classifier & \textbf{0.9300} (0.0163) & \textbf{0.9320} (0.0155) & \textbf{0.9300} (0.0163) & \textbf{0.9300} (0.0161) \\
\hline
\end{tabular}
}

\end{table}

\begin{table}[h]
\centering
\caption{Accuracy of proposed AdaptMoist model on wood chip datasets from different sources and batches based on different texture features. S1$\to$S2 indicates training the models on S1 and testing on S2.}
\vspace{0.3cm}
\label{tab:da_models}
\resizebox{\linewidth}{!}{
\begin{tabular}{c|cccccc|cc|c}
\hline
Models                                                        & S1$\to$S2 & S1$\to$S3 & S2$\to$S1 & S2$\to$S3 & S3$\to$S1 & S3$\to$S2 & S2B1$\to$S2B2 & S1B2$\to$S2B1 & Avg.\\ \hline

Haralick &    \textbf{0.80}       &     \textbf{0.86}      &     \textbf{0.72}      &    \textbf{0.76}       &    \textbf{0.74}       &   \textbf{0.78}        &    0.87           &      0.84        & \textbf{0.80} \\ 
FOS                                                & 0.60 & 0.43 & 0.54 & 0.39 & 0.51 & 0.45 & 0.80 & 0.77 & 0.56\\
FPS                                               & 0.38 & 0.30 & 0.46 & 0.40 & 0.40 & 0.30 & 0.68 & 0.66 & 0.60\\
GLRLM  & 0.43 & 0.30 & 0.55 & 0.51 & 0.57 & 0.30 & 0.69 & 0.77 & 0.52\\
LBP                                                 & 0.34 & 0.30 & 0.39 & 0.42 & 0.28 & 0.65 & 0.69 & 0.71 & 0.47 \\
Combined                                                & 0.53 & 0.40 & 0.56 & 0.43 & 0.40 & 0.30 & \textbf{0.91} & \textbf{0.85} & 0.55\\ \hline
\end{tabular}
}
\end{table}

\clearpage
\section*{CRediT Author Statement}
\textbf{Abdur Rahman:} Conceptualization, Methodology, Software, Formal analysis, Data Curation, Writing - Original Draft, Visualization. \textbf{Jason Street:} Conceptualization, Data Curation, Writing - Review \& Editing, Funding acquisition. \textbf{Mohammad Marufuzzaman:} Conceptualization, Writing - Review \& Editing, Supervision, Funding acquisition. \textbf{Haifeng Wang:} Conceptualization, Methodology, Writing - Review \& Editing, Funding acquisition. \textbf{Veera G. Gude:} Conceptualization, Funding acquisition. \textbf{Randy Buchanan:} Conceptualization, Funding acquisition.

\section*{Declaration of competing interest}
The authors declare that they have no known competing financial interests or personal relationships that could have appeared to influence the work reported in this paper.

\section*{Acknowledgements}
This work is supported by the Sustainable Bioeconomy through Biobased Products and Engineering for Agricultural Production and Processing programs, project award no. 2020-67019-30772 and 2022-67022-37861, from the U.S. Department of Agriculture’s National Institute of Food and Agriculture. Any opinions, findings, conclusions, or recommendations expressed in this publication are those of the author(s) and should not be construed to represent any official USDA or U.S. Government determination or policy.
 %\bibliographystyle{vancouver}
 \bibliographystyle{elsarticle-num}
 \bibliography{cas-refs}





% \end{thebibliography}
\end{document}